\documentclass[a4paper,12pt]{extarticle}
\usepackage{graphicx}
\usepackage{pdfpages}
\usepackage{multido}

\usepackage{tikz}
\usepackage{comment}
\usepackage{graphicx}
\usepackage{amsmath}
\usepackage{amssymb}
\usepackage{booktabs}
\usepackage{calc}
\usepackage{mathtools}
\usepackage{multirow} 

\usepackage{geometry}
\usepackage{titling}
\usepackage{caption}
\usepackage{subcaption}

\usepackage[pagebackref,breaklinks, colorlinks]{hyperref}

\usepackage[capitalize]{cleveref}
\crefname{section}{Sec.}{Secs.}
\Crefname{section}{Section}{Sections}
\Crefname{table}{Table}{Tables}
\crefname{table}{Tab.}{Tabs.}

\usepackage[accsupp]{axessibility}
\usepackage{array}
\newcolumntype{L}[1]{>{\arraybackslash}p{#1}}

\def\ie{\emph{i.e. }}

\newcommand\blfootnote[1]{%
  \begingroup
  \renewcommand\thefootnote{}\footnote{#1}%
  \addtocounter{footnote}{-1}%
  \endgroup
}

\definecolor{somegray}{rgb}{0.5, 0.5, 0.5}
\newcommand{\darkgrayed}[1]{\textcolor{somegray}{#1}}
\makeatletter
\newcommand*\titleheader[1]{\gdef\@titleheader{#1}}
\AtBeginDocument{%
  \let\st@red@title\@title
  \def\@title{%
    \vskip-3em
    \bgroup\normalfont\large\centering\@titleheader\par\egroup
    \vskip1.5em\st@red@title}
}
\makeatother

\renewcommand\maketitlehooka{%
  \setlength\parindent{0pt}%
  \begin{minipage}{\textwidth}
    \begin{minipage}{\textwidth}
        \begin{center}
        \darkgrayed{This paper has been accepted for publication at the \\
        European Conference on Computer Vision (ECCV), Tel Aviv, 2022.
        \copyright Springer}
        \end{center}
    \end{minipage}%
  \end{minipage}%\vskip 2.5ex
}

\begin{document}

\title{\includegraphics[scale=0.04]{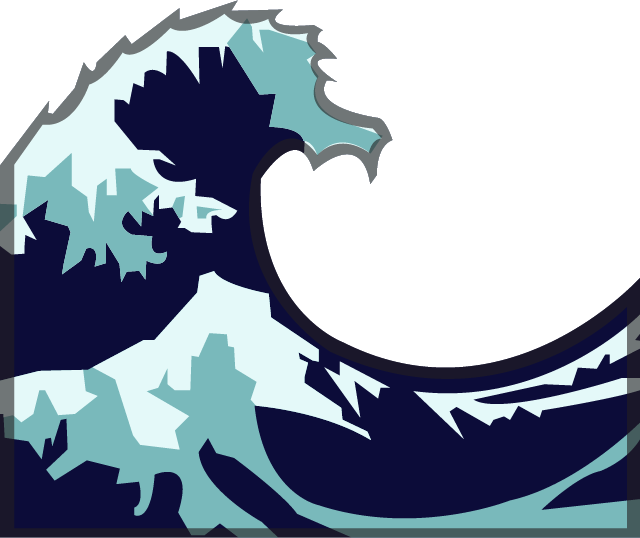} \textbf{Online Domain Adaptation for Semantic Segmentation in Ever-Changing Conditions} }

\author{Theodoros Panagiotakopoulos$^{1*}$\quad\quad
Pier Luigi Dovesi$^2$ \\
Linus H{\"a}renstam-Nielsen$^{3,4 *}$ \quad\quad
Matteo Poggi$^5$
\\
%\authorrunning{T. Panagiotakopoulos et al.}
\small $^1$King \quad\quad $^2$Univrses \quad\quad
$^3$Kudan \quad\quad $^4$ Technical University of Munich \\ \small $^5$University of Bologna
}
%******************
\date{ }

{
\vspace{-2cm}
\maketitle
\begin{center}
    \vspace{-2cm}
    \begin{tabular}{c c c c}
        \includegraphics[width=\linewidth]{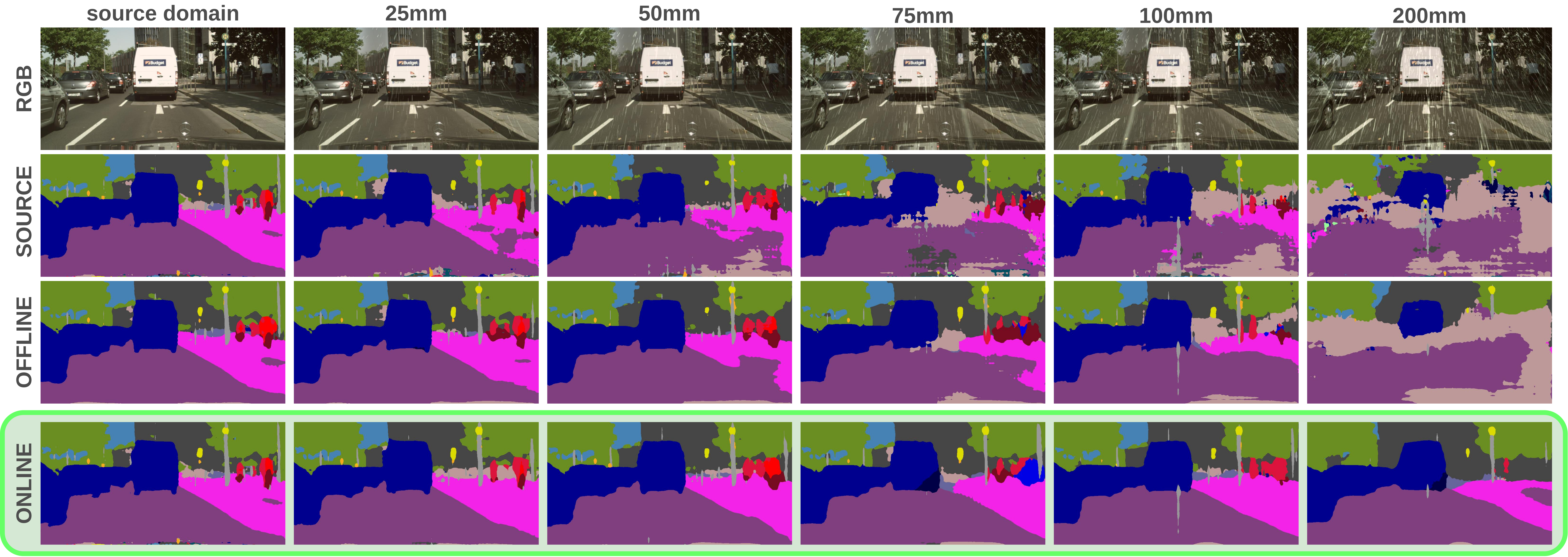} \\
    \end{tabular}
    \label{fig:teaser}
\end{center}
\vspace{-0.2cm}
\small \hypertarget{fig:teaser}{Figure 1.} \textbf{OnDA framework in action.} We show images with varying intensity of rain (from 0 to $200$mm). When dealing with such complicated domain shifts, both pretrained networks and offline adaptations struggle, whereas our online framework is able to adapt, without forgetting.
%\vspace{-0.5cm}
}

%%%%%%%%% ABSTRACT
\begin{abstract}
Unsupervised Domain Adaptation (UDA) aims at reducing the domain gap between training and testing data and is, in most cases, carried out in offline manner.
However, domain changes may occur continuously and unpredictably during deployment (e.g. sudden weather changes). In such conditions, deep neural networks  
witness dramatic drops in accuracy and offline adaptation may not be enough to contrast it.
In this paper, we tackle Online Domain Adaptation (OnDA) for semantic segmentation. We design a pipeline that is robust to continuous domain shifts, either gradual or sudden, and we evaluate it in the case of rainy and foggy scenarios.
Our experiments show that our framework can effectively adapt to new domains during deployment, while not being affected by catastrophic forgetting of the previous domains.
\end{abstract}

%%%%%%%%% BODY TEXT
\section{Introduction}
\label{sec:intro}

\blfootnote{$^*$ Part of the work carried out while at Univrses.}The task of semantic segmentation consist of assigning each pixel of an image to a specific class. With the spread of deep learning, Convolutional Neural Networks (CNNs) have been established as the state-of-the-art for tackling this kind of problem \cite{chen2017deeplab,yuan2021segmentation,chen2020naive}. However, despite training on a large quantity of annotated images, the network predictions can often be unreliable when deployed on new scenarios, because of the \textit{domain shift} occurring between training and deployment. For example, the shift can be due to the images being collected in very different environments (e.g., urban versus rural roads) or lighting conditions (e.g., day versus night).

Consequently, Unsupervised Domain Adaptation (UDA) arose as a popular research trend to overcome the domain shift problem. It aims at shrinking the gap between a labeled set of images -- the \textit{source} domain, over which supervised training is possible -- and an unlabeled one -- the \textit{target} domain, for which ground truth annotations are not available. This is performed in several ways, such as transferring the image style across the two \cite{cyclegan,dcan,cycada}, or either conditioning or normalizing the feature space \cite{ganin,fada}. Techniques for UDA have been extensively studied in the offline setting, thus assuming to have availability of both the source and the target domain images in advance, then proceeding by adapting a model, trained with ground truth supervision on source, to the target. However, such an assumption is often too strong to hold in the context of an actual application. 
We argue that domain shifts are likely to continuously arise during deployment. Some examples can be related to different cities or weather conditions, or even, at a lower level, involving different camera positioning and intrinsics.

While some domain shifts come in predictable ways (e.g., day-night cycle), some others can occur unpredictably -- such as weather changes, either in a slow (e.g., rain, fog) or sudden way (e.g., storms).
Leveraging the fact that environmental changes may often happen gradually, we propose an online adaptation pipeline that exploits progressive adaptation. 
Inspired by recent progress in Curriculum Learning applied to UDA, we expand the paradigm beyond the adaptation to an \textit{intermediate} domain by designing a framework able to autonomously identify domain changes and adapt its self-training policy accordingly. In online settings, we seek to seamlessly find the optimal response to the current deployment domain while, crucially, we would like to proactively prepare for future scenarios. We argue that online settings need to break the dichotomy between Source and Target domain, where the Target has now to be modeled as a ``domain sequence''. Extensive empirical studies will highlight how the good modeling of the domain sequence is paramount for this purpose. Indeed, most of the improvements observed in specific target domains are gained ``in advance'', i.e. before the model has ever been exposed to that specific distribution.

To perform online adaptation, we benchmark our model on increasing intensities of rain ($25$, $50$, $75$, $100$, and $200$mm of rain) and fog ($750$, $375$, $150$, and $75$m of visibility). We demonstrate that deep learning models aware of domain shift \textit{intensity} and \textit{direction} can exploit intermediate domains substantially better.
We achieve this by introducing an active teacher-model switching mechanism that allows for higher adaptation flexibility, hence reaching farther target domains, as visible in Figure \ref{fig:teaser}. Additionally, when needed, the switching mechanism can revert the adaptation process allowing adapting back to source domain without experiencing catastrophic forgetting.
Our main contributions are:

\begin{itemize}
    \item We introduce an online progressive adaptation benchmark for UDA methods.
    \item We propose an approach that leverages progressive adaptation to increase performance on distant domains in an online manner.
    \item We demonstrate that catastrophic forgetting can be avoided by actively updating the self-training policy during adaptation and using a Replay Buffer.
    \item We run experiments on various simulated scenarios and, crucially, we show that models that have been previously exposed to gradual domain adaptation can acquire the ability to cope with sharp changes as well.
\end{itemize}

\section{Related Work}
Online Domain Adaptation is directly connected to many fields of Machine Learning, such as Transfer Learning and Continuous Learning. We now review methods that focus on reducing the domain gap, lessening the effect of catastrophic forgetting, and continuously adapting to upcoming domains.

\textbf{Unsupervised Domain Adaptation.} Unsupervised Domain Adaptation (UDA), is a relatively new field that has gained interest due to the rising amount of data and the limited and expensive resources needed to annotate them. 
Early UDA approaches focus on constructing domain invariant feature representations \cite{ganin} or transferring the ``style'' from one domain to another, for instance, by means of the CycleGAN \cite{cyclegan} framework.
First attempts \cite{dcan,cycada} learned this transfer in an offline manner before training, then translating images during training itself. More modern approaches \cite{bdl,stylization,yang_fda_2020} combine the two phases in one in an end-to-end framework. This strategy has been extended by LTIR \cite{ltir}, in order to learn texture-invariant features by training on source images augmented with textures coming from other real images.
Often, adversarial learning has been deployed for UDA aiming at obtaining better alignment of the source and target distributions, either in features \cite{ganin,fada,chen_no_2017,hoffman_fcns_2016} or output \cite{adaptsegnet} spaces. Later works \cite{cicek_unsupervised_2019,chen_no_2017} highlight the use of class information in adversarial learning, while Advent \cite{advent} introduces an adversarial approach to perform entropy minimization.

\textbf{Self-training.}
Recent trends concerning UDA leverage the idea of producing \textit{pseudo-labels} \cite{pseudolabel} for self-training over the target domain, inspired by the recent success in semi-supervised tasks \cite{zhai2019s4l,sohn_fixmatch_nodate}. Since these labels are noisy, designing robust strategies to reduce the effect of wrong labels is of paramount importance for this family of approaches. 
\cite{cbst} implements this by means of a confidence-based thresholding algorithm, \cite{iast} extends this approach with an instance adaptive variant, further improving the quality of the produced pseudo-labels. 
Nevertheless, naive pseudo-labeling can produce unreliable confidence estimates and an increased bias towards the most common classes. To contrast this \cite{zou2018unsupervised,hoyer2021daformer} propose approaches that balances class predictions, while \cite{zou2019confidence} regularizes the model confidence.
On this same track, \cite{mrnet} uses pseudo-labels to minimize the discrepancy between two classifiers, while \cite{Pan_2020} aims at inter-domain and intra-domain gap minimization, supported by pseudo-labels, and \cite{Cardace_2022_WACV} uses shallow features to improve class boundaries.
Finally, newer approaches \cite{chen2019progressive,zhang_category_2019,zhang_prototypical_2021} leverage \textit{prototypes}, defined as feature-space class centroids, to produce unbiased pseudo-labels.

\textbf{Source-free UDA} or ``model adaptation'', is a topic that was introduced to assist continual learning \cite{Stan2021UnsupervisedMA}.  In contrast to traditional unsupervised domain adaptation, the use of source and target samples happens separately. Therefore, the learning approach consists of two separate steps, the \textit{task learning step} using the source data and the \textit{adaptation step} using the target. Several approaches have been explored: \cite{liu_source-free_2021} tries to solve the lack of source samples by deploying a generator that produces samples that resemble the source data. In contrast, \cite{liang_we_2020} freezes the final layers of the network and performs self-training. Similarly \cite{wang2021tent} retrains Batch-Normalization layers through entropy minimization. To avoid forgetting source during adaptation, \cite{liu2021ttt} introduces a feature alignment during adaptation. Finally, \cite{iwasawa2021testtime} uses the distance between embeddings and test-time adapting prototypes to compute the predictions.
Notably, these latter \cite{liu2021ttt,iwasawa2021testtime} have been proposed and tested for classification tasks on toy datasets.

\textbf{Curriculum Learning} is a training strategy that focuses on the order in which information is exploited. As described by \cite{bengio_curriculum}, machine learning models can learn much better when information is presented in a meaningful order.
\cite{liu_open_2020} developed a domain encoder to express domain distance. The target domains were then ordered based on their similarity to the source domain. Adaptation was then performed from the closest to the furthest from the source domain. \cite{sakaridis_model_2018} propose using generated foggy images as an intermediate step to adapting to real weather scenarios.

\textbf{Continuous UDA.}
Some works tried to integrate UDA with continual learning, tackling the problem of ``adapting without forgetting''. Several methods employ Replay Buffers \cite{bobu_adapting_2018,lao2020continuous,kuznietsov2022towards}, ACE \cite{wu_ace_2019} leverages AdaIN \cite{huang2017arbitrary} to perform style transfer while retaining previous knowledge through a task memory. \cite{su_gradient_2020} adapts through Contrastive Learning while constraining the gradient to reduce forgetting, and \cite{wulfmeier_incremental_2018} uses a generator to produce the necessary data to perform adversarial training. 

Despite the large body of existing literature, as raised by a contemporary work \cite{shift}, current datasets and UDA methods fall short of representing and testing on realistic online scenarios, \ie{with incremental domain-shifts occurring continuously with the flowing of input images}.

\section{Online Domain Adaptation}
\label{sec:methods}
\begin{figure}[t]
    \centering
    \includegraphics[width=0.78\textwidth]{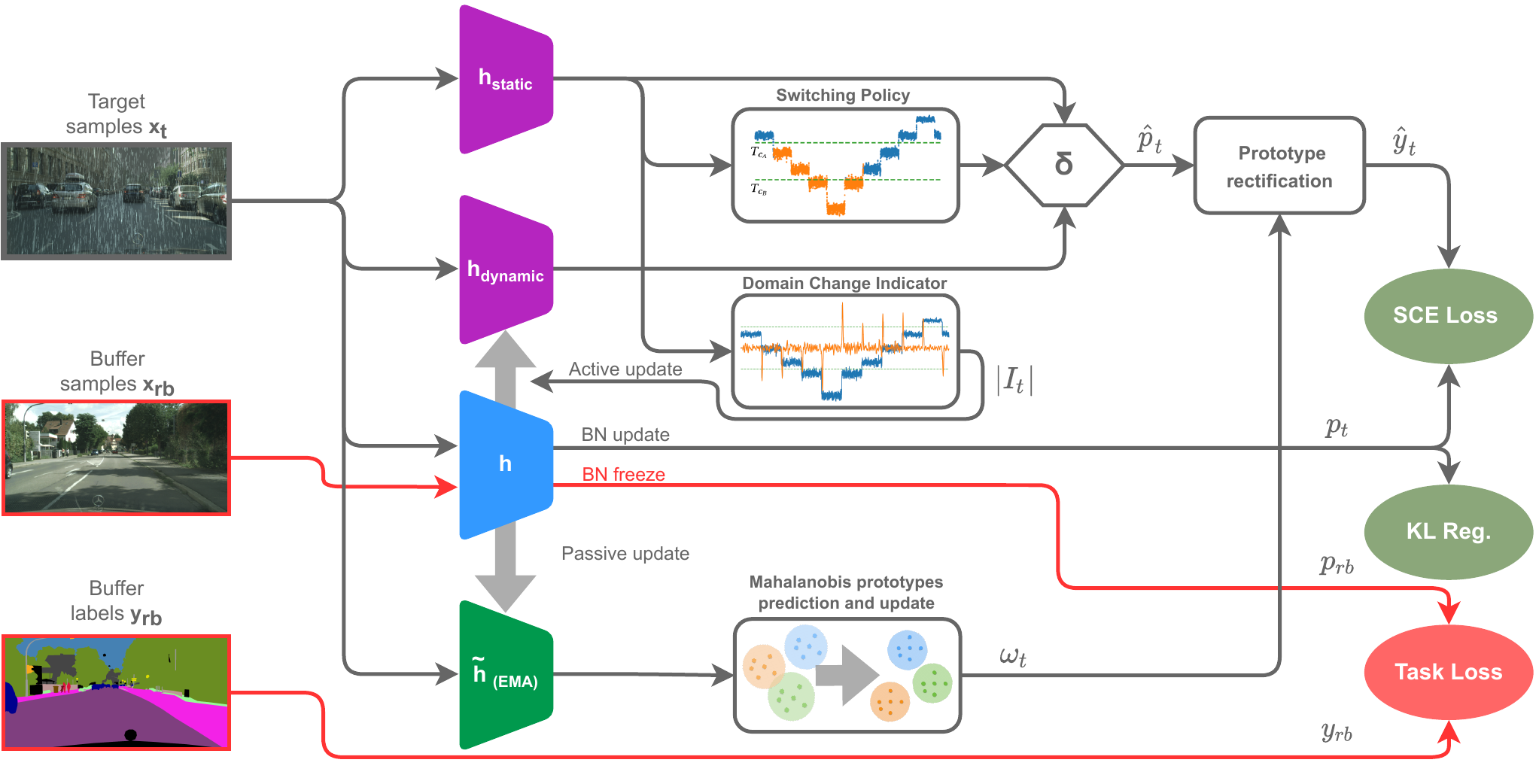}
    \caption{\textbf{Overview of our OnDA framework.} It comprises Switching Policy, Domain Change Indicator, Mahalanobis prototypes prediction, and BN freezing. }
    \label{fig:flowchart}
\end{figure}

This section introduces our framework for Online Domain Adaptation (OnDA) specific to face ever-changing environments. While adopting state-of-the-art UDA strategies for prototypical self-training \cite{zhang_prototypical_2021,zou2019confidence,wang_symmetric_2019}, we design a novel strategy to address online settings. We present a student-teacher approach \cite{furlanello2018born} which allows for dynamic teachers orchestration by both actively updating the teacher according to the \textit{domain change} and by strategically choosing the best teacher to employ to train the student model. Furthermore, we propose to exploit feature variance for better prototype predictions, we investigate the impact of BatchNorm during adaptation and we assess the importance of a Replay Buffer to prevent catastrophic forgetting. We present an overview of OnDA in Figure \ref{fig:flowchart}. 

%  ---- intro: ----

\subsection{Online Prototypical Self-Training}
\label{sec:3_prototypical_pseudolable}

We now introduce the design of a prototypical framework for the online setting. 

\textbf{Replay Buffer.} First of all, in order to simulate a realistic deployment, where storing the full source dataset $D_S$ might be infeasible, we sample a subset $D_{RB}\subseteq D_S$ as a Replay Buffer. The buffer is used for training with segmentation loss during adaptation and prevents the network from forgetting the original domain. As we will show in Section \ref{sec:ablation}, even a small buffer is very helpful in the mitigation of catastrophic forgetting.

\textbf{Problem formulation.}
We define our network as $h = g \circ f$, where $f$ is the feature encoder mapping images into a space of dimension $K$ and $g$ maps features into class labels. We denote sample-label pair from the source and Replay Buffer as ($x_s,y_s$) and ($x_{rb},y_{rb}$) respectively.
We model the target domain $D_{T}$ as a sequence of $\Theta$ sub-domains, such that $D_{T}=(D_{T_1}, D_{T_2}, ..., D_{T_\Theta})$.

\textbf{Prototypes initialization.}  In online scenarios target samples appear sequentially ($x_t^{(1)}, x_t^{(2)}, ...,x_t^{(N)}$) and it is unknown from which $D_{T_\theta}$ they have been sampled. Therefore, target samples can not be used to initialize the class prototypes before the adaptation process takes place. To address this limitation, we initialize the prototypes using the source dataset and update them on the fly using target samples. Letting $h_\text{static} = g_\text{static} \circ f_\text{static}$ be the network fully trained on $D_S$ before adaptation, the prototype initialization  $\eta^c \in \mathbb{R}^K$ for each class $c$ is given by:
\begin{equation}
\eta^{c}=\frac{1}{\left|\Lambda_{\mathbf{S}}^{c}\right|} \sum_{i=1}^{N_S} \sum_{j}^{H \times W} \left(y_s^{(i)}|_{j} = c \right) \left.f_\text{static}\left(x^{(i)}_{s}\right)\right|_{j}
\label{eq:3_prototypes_init}
\end{equation}
where $N_S$, $H$, $W$ are the number of source samples, image height and image width respectively. $\left|\Lambda_{\mathbf{S}}^{c}\right|$ denotes the number of pixels that belong to class $c$ from set S, and $\left|\Lambda_{\mathbf{S}}\right| = \sum_c \left|\Lambda^c_{\mathbf{S}}\right|$. The variance $\sigma \in \mathbb{R}^K$ of each dimension $k$ of the prototype space is then obtained through:
\begin{equation}
\begin{aligned}
\sigma^2=\frac{1}{\left|\Lambda_{\mathbf{S}}\right|} \sum_{i=1}^{N_S} \sum_{j}^{H \times W} \left.{f_\text{static}\left(x^{(i)}_{s}\right)}^2\right|_{j}
- \left(\frac{1}{\left|\Lambda_{\mathbf{S}}\right|} \sum_{i=1}^{N_S} \sum_{j}^{H \times W} \left.f_\text{static}\left(x^{(i)}_{s}\right)\right|_{j}\right)^2&.
\end{aligned}
\label{eq:3_proto_var}
\end{equation}
Hence, given a target sample $x_t$, we obtain the prediction $\omega_t^{c}$ as the softmax of the variance-normalized proximity between prototypes $\eta$ and the momentum encoder prediction $\tilde{f}\left(x_{t}\right)$,

\begin{equation}
\omega_{t}^{c}=\frac{\exp \left(-\left\|\left(\tilde{f}\left(x_{t}\right)-\eta^{c}\right) / \sigma\right\|\right)}{\sum_{c^{\prime}} \exp \left(-\left\|\left(\tilde{f}\left(x_{t}\right)-\eta^{c^{\prime}}\right) / \sigma\right\|\right)}
\label{eq:3_proto_softmax_pred}
\end{equation}
where $ \tilde{h} = \tilde{g} \circ \tilde{f} $ is the momentum model of the network $h$. The momentum encoder $\tilde{f}$ is used to produce stable predictions compared to using the main encoder $f$ directly and it is created by exponentially averaging the parameters of $f$ over time. In Eq. \ref{eq:3_proto_softmax_pred}, we employ a form of Mahalanobis distance, including the variances in each dimension of the feature vector. Compared to the Euclidean distance, this leads to a more accurate measure of the distance and higher overall metrics (+4.5\% mIoU in the hardest target domain). 

\textbf{Prototypes update.} During the adaptation, the prototypes are updated online through target samples pseudo-labeling. Given a batch of $N$ target samples $B$, the batch prototypes are defined as follows:

\begin{equation}
\hat{\eta}^{c}=\frac{1}{\left|\Lambda_{\mathbf{B}}^{c}\right|} \sum_{i=1}^{N} \sum_{j}^{H \times W} \left(\tilde{h}(x_t^{(i)}) = c \right) \left.\tilde{f}\left(x^{(i)}_{t}\right)\right|_{j} .
\label{eq:3_prototypes_update}
\end{equation}
\noindent
Then for all classes $c$ where $\left|\Lambda_{\mathbf{B}}^{c}\right| > 0$ we update the  corresponding prototype using $\eta^c \longleftarrow \lambda \eta^c + (1-\lambda) \hat{\eta}^c$. Prototypes are used to lessen the reliance on the source label distribution. Naive pseudo-labeling will produce models that are highly biased towards the most popular and easier classes. In domain adaptation, source and target label distributions do not necessarily align. Feature distance through prototypes, on the other hand, removes class biases and produces unbiased predictions. Finally, the pseudo-label $\hat{y}_t$ for a sample $x_t$ is computed by rectifying the model prediction using the prototype softmax output $\omega_{t}$ as follows:
\begin{equation}
   \hat{y}_{t} = \xi(\hat{p}_{t} \cdot \omega_{t})
   \label{eq:3_pseudolabeling}
\end{equation}
where $\xi$ is a function that transforms soft-labels to one-hot encoded hard labels. 
Moreover, instead of directly using the model prediction $p_t= h(x_t)$, checkpoints of the $h$ model are used. In particular, we define:
\begin{equation}
    \hat{p}_{t} = \delta h_{\text{static}}(x_t) + (1-\delta)  h_{\text{dynamic}}(x_t)
    \label{eq:3_prior_weighted_avg}
\end{equation}
Where $h_\text{dynamic}$ is the last adapted model on the previous deployment domain, and $\delta \in [0, 1]$ determines the contribution of each model. In Section \ref{sec:3_prior_switching} we will describe how to guide the adaptation process by dynamically updating $\delta$.

\textbf{Overconfidence handling.} Self-training is a form of entropy minimization which means that the network will tend to become overconfident. Pseudo-labeling with thresholding strategies alone fail since confidence is no longer a reliable guideline. We utilize loss functions that can withstand overfitting to noisy labels. For this reason, two key techniques are used: confidence regularization and Symmetrical cross-entropy \cite{wang_symmetric_2019}. Given the model prediction $p_t = h(x_t)$ we apply a KL divergence regularizer \cite{zou2019confidence}

\begin{equation}
\mathcal{L}_\text{reg} = 
-\gamma\sum_{k=1}^{K} \frac{1}{K} \log p_t .
\label{eq:3_confidence_reg}
\end{equation}
Moreover, to mitigate the impact of noisy labels we employ Symmetrical Cross-Entropy (SCE). The pseudo-label loss is then described as follows:
\begin{equation}
    \mathcal{L}_\text{pseudo} = \alpha \ell_\text{ce}\left(p_{t}, \hat{y}_{t}\right)+\beta \ell_\text{ce}\left(\hat{y}_{t}, p_{t}\right) .
\end{equation}
\noindent
Where $\ell_\text{ce}$ is the Cross-Entropy loss, and $\alpha$ and $\beta$ are two weighting hyper-parameters.
The complete loss function to perform learning using source and target samples is defined as follows:
\begin{equation}
    \mathcal{L}_\text{total} = \mathcal{L}_\text{task}(x_{rb}, y_{rb}) + \mathcal{L}_\text{pseudo} (x_t, \hat{y}_{t}) + \mathcal{L}_\text{reg}(x_t) .
\end{equation}

\textbf{Batch normalization switching.} Batch Normalization (BN) layers \cite{ioffe2015batch} are employed to normalize features so as to obtain zero-mean and unit standard deviation distributions by iteratively accumulating statistics after processing any batch. Given features $x_i$ for the $i$-th element of the batch, the output $y_i$ of any BN layer is computed as $y_i = \frac{x_i-\mu_\text{B}}{\sigma_\text{B}}$ with $\mu_\text{B},\sigma_\text{B}$ being the exponential moving average of the mean and variance of features $x$ respectively.
During online adaptation, our network processes data from two different distributions; the samples in the Replay Buffer ($\mathcal{D}_\text{RB}$) and those belonging to the target domain distribution ($\mathcal{D}_{T_{\theta}}$). This leads to cumulative statistics not being meaningful as an actual distribution, as already observed in \cite{chang_domain-specific_2019,klingner2022unsupervised}.
Hence, we investigate two approaches to batch normalization: i) freezing BN layers when processing samples from $\mathcal{D}_\text{RB}$ or ii) swapping BN statistics between $\mathcal{D}_\text{RB}$ and $\mathcal{D}_{T_x}$. Both turn out to be beneficial in our online settings, thus we selected i) for simplicity. We provide a comparison between the BN approaches in the supplementary material. 

\subsection{Domain Shift Detection}
\begin{comment}
\begin{figure}[t]
    \centering
    \begin{subfigure}[h]{0.4\textwidth}
    \includegraphics[width=\textwidth]{images/switching/confidence_derivative_rasterised.pdf}
    \end{subfigure}
    %\vspace{-0.2cm}
    \caption{\textbf{Working principle of the domain switching detector.} On the back, in blue, the $h_{\text{static}}$ confidence is displayed, while with orange color the derivative $\mu_{z_{t}} - \mu_{z_{t-1}}$ is showed. When switching to more distant domains the value function has negative peeks and a positive on closer domains.
    }
    \label{fig:3_derivative}
    %\vspace{-0.5cm}
\end{figure}
\end{comment}
\label{sec:3_indicator}
An essential part of online adaptation is being able to detect the domain shifts and act accordingly. Inspired by the key role of the model confidence as a mean for measuring and minimizing domain shift \cite{vu_advent_2019,vu2019dada}, we use the confidence of $h_\text{static}$ to identify domain changes as well as the ``direction'' of such a change. In online settings, it is indeed crucial to both recognize whether the deployment domain is changing and if the transition is leading towards a more distant domain (\textit{forward}) or a closer domain (\textit{backwards}), relatively to the source.
Defining $z_t$ as the confidence of $h_\text{static}$ over the $t$-th batch, expressed as
\begin{equation}
z_t=\frac{1}{\left|\Lambda_{\mathbf{B}}\right|} \sum_{i=1}^{N} \sum_{j}^{H \times W} \max_c \left.h_{\text{static}}\left(x^{(i)}\right)\right|_{j,c}
\label{eq:3_confidence}
\end{equation}
we notice clear changes while transitioning between domains (see Figure \ref{fig:flowchart}, the Domain Change Indicator, and Sec. 3 of the supplementary material).
Therefore, the confidence derivative can be leveraged as an indicator of domain changes and computed using the difference between consecutive values. In order to reduce noise and have a robust representation, a shifting window of length $n$ is used. On each new batch $t$, confidence values ($z_t$) are appended to the window, while old ones are removed ($z_{t-n}$). At any given time we compute the weighted average confidence of the window as $\mu_{t} = \frac{1}{n}\sum_{i=0}^n w[i]z_{t -i}$, where $w$ is the discrete Hamming window \cite{poularikas_handbook_1999} of length $n$. The switching indicator function can then be defined as:
\begin{equation}
    I_t = \left\{
\begin{matrix*}[l]
\;\;\;1, \;\;\;\;\; \mu_{t} - \mu_{t-1} > T_\text{cd}
\\ -1, \;\;\;\;\; \mu_{t} - \mu_{t-1} < -T_\text{cd}
\\ \;\;\;0, \;\;\;\;\; \text{otherwise} 
\end{matrix*}\right., \; \forall \; t > n+1 .
\end{equation}
When the window has not been filled yet ($t \leq n+1$), we set $I_t=0$. $T_\text{cd}$ is a hyperparameter that controls model sensitivity to domain changes. We then detect domain changes by examining the absolute value $|I_t| > 0$ and their direction studying its sign.
In the supplementary material video, we present the behavior of the Domain Change Indicator for much more challenging scenarios, which led us to introduce a simple debouncing window to ensure robust switching.

\subsection{Prior model Switching techniques}
\label{sec:3_prior_switching}
In this section we will focus on the prior predictions $\hat{p}_{}$ introduced in Eq. \ref{eq:3_prior_weighted_avg}. In particular, two models are used to acquire the prior: $h_\text{static}$ and $h_\text{dynamic}$. $h_\text{static}$ is the initial model, before any adaptation, while $h_\text{dynamic}$ is the model before the adaptation to the current domain takes place. For example, in an adaptation sequence over domains $D_{T_1}, D_{T_2}, D_{T_3}$, during the first adaptation ($D_{T_1}$) the dynamic and static models coincide. As the switching indicator $I_t$ perceives that we are moving to the second domain ($D_{T_2}$), $h_\text{dynamic}$ is updated becoming the model right after the adaptation on $D_{T_1}$. Similarly, for the third adaptation (on $D_{T_3}$), $h_\text{dynamic}$ will become the model after the adaptation on $D_{T_2}$ and before $D_{T_3}$. The choice of which teacher model is going to be used for the prototypes rectification heavily influence the adaptation capabilities.

\begin{figure}[t]
    \centering
    \begin{subfigure}[t]{0.24\textwidth}
    \centering
    \includegraphics[width=1\textwidth]{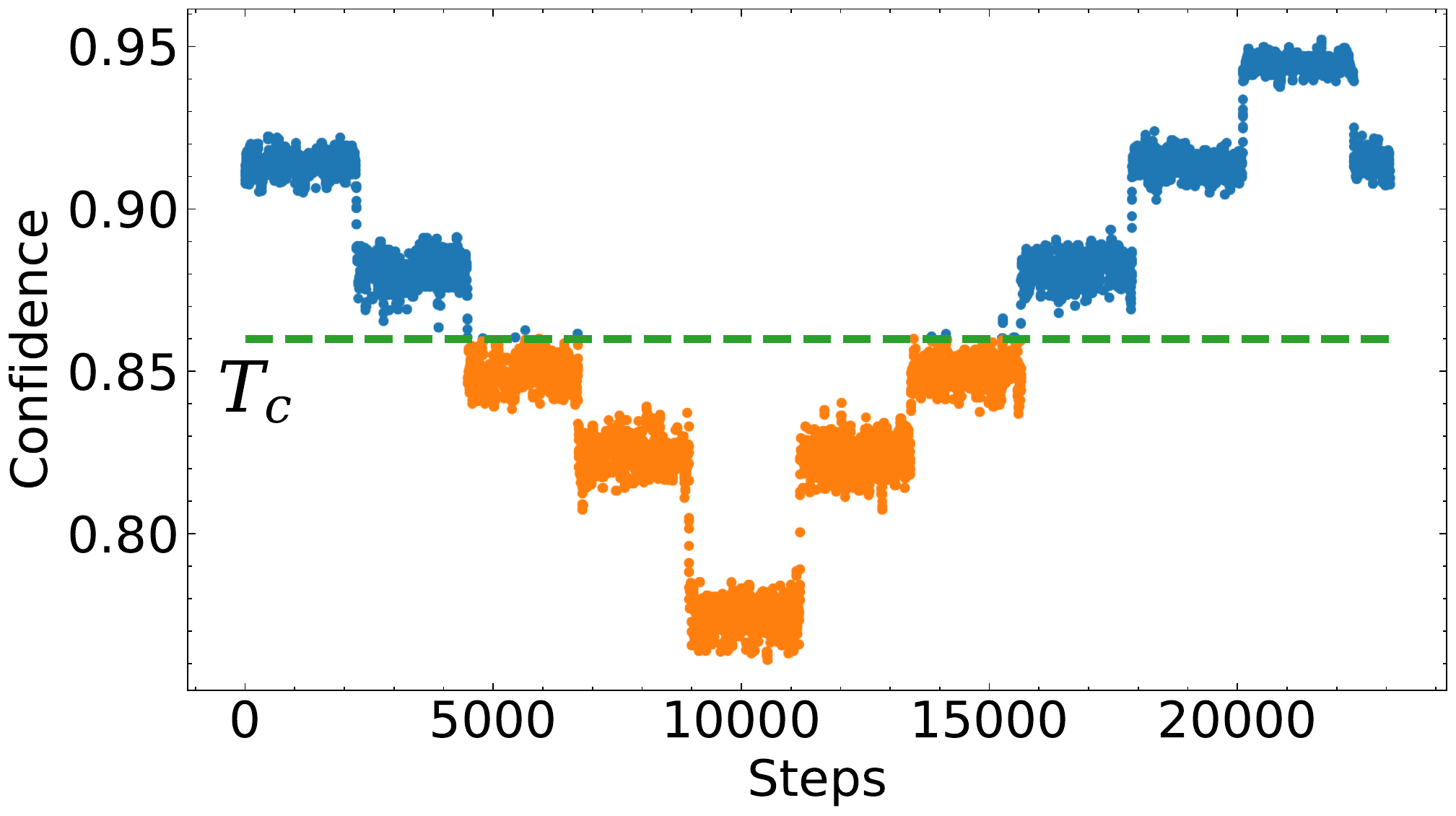} \\
    \footnotesize{Confidence Switch}\\
    \end{subfigure}
    \begin{subfigure}[t]{0.24\textwidth}
    \centering
    \includegraphics[width=1\textwidth]{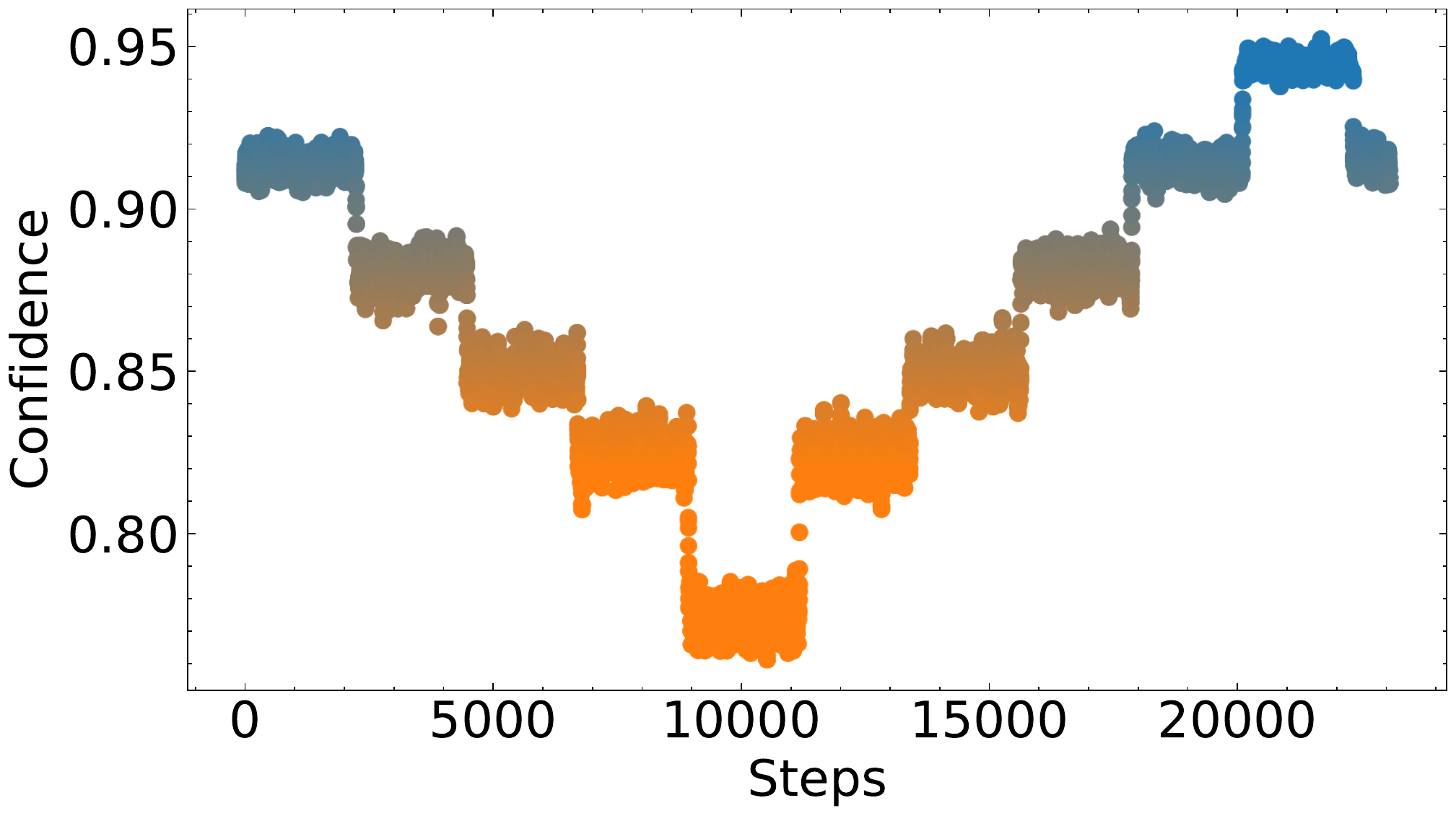} \\
    \footnotesize{Soft-Confidence Switch}
    \end{subfigure}
    \begin{subfigure}[t]{0.24\textwidth}
    \centering
    \includegraphics[width=1\textwidth]{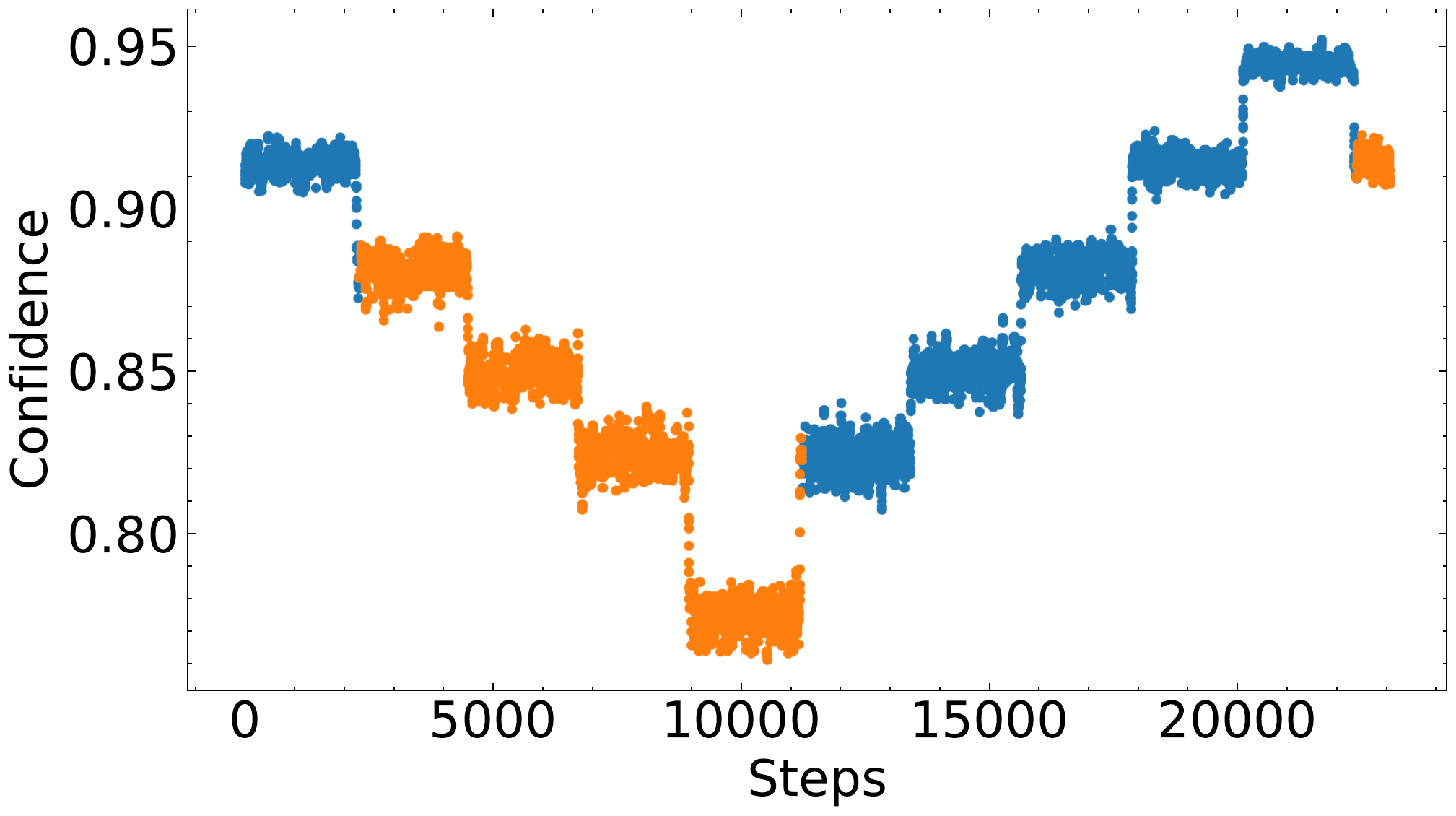}\\
    \footnotesize{Confidence Derivative Switch}\\
    \end{subfigure}
    \begin{subfigure}[t]{0.24\textwidth}
    \centering
    \includegraphics[width=1\textwidth]{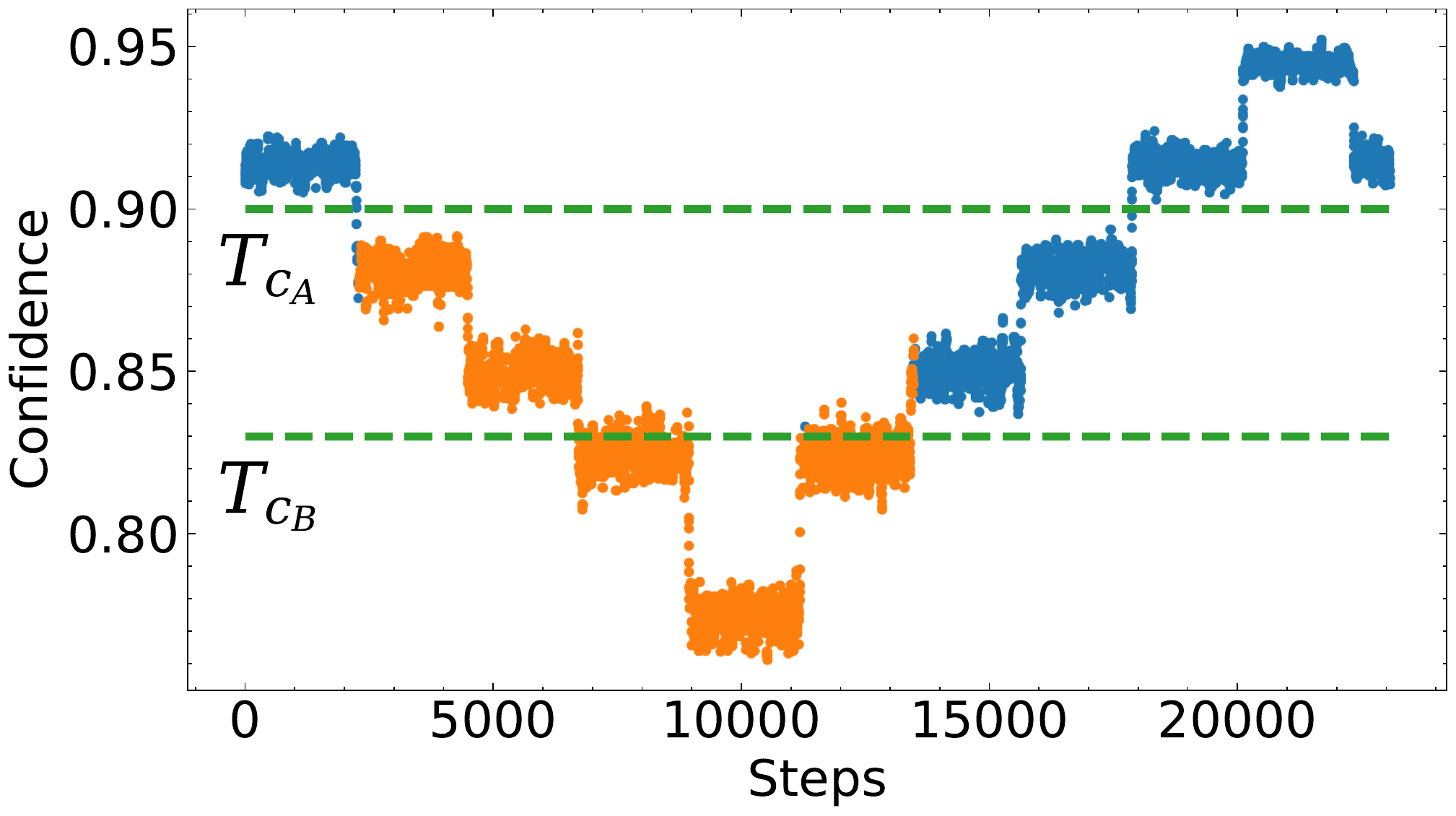}\\
    \footnotesize{Hybrid Switch}
    \end{subfigure}
    \caption{\textbf{Visualization of the switching policies.} The dots show the static models confidence values over time and their color represents $\delta$ values: blue corresponds to $\delta=1$, i.e. $h_\text{static}$, while orange $\delta=0$, i.e. $h_\text{dynamic}$. The \textit{Soft-Confidence Switch} performs a linear transition from one prior to the other and it is represented through a color gradient. 
    }
    \label{fig:3_switching}
\end{figure}

In the experimental section, we will show that employing $h_\text{dynamic}$ ($\delta=0$) grants extra flexibility, allowing adaptation to \textit{harder} (i.e. more distant) domains. On one hand, as a drawback, it performs sub-optimally while ``adapting back'' to previous domains and even $D_S$.  On the other hand, $h_\text{static}$ ($\delta=1$) intrinsically limits adaptation due to its predictions guiding the self-training towards the original model. Nonetheless, it allows for better adaptation in domains closer to $D_S$, preventing catastrophic forgetting. We thus identify the need for a mechanism that allows for an effective switching between the two prior models, hence making $\delta$ function of $\mu_t$, i.e. $\delta_t = Switch(\mu_t, t)$. We introduce the following policies, summarized in Figure \ref{fig:3_switching}:

\begin{itemize}
    \item \textit{Confidence Switch (CS)}: Applies simple thresholding on the static model confidence $z_t$.
    \begin{equation}
    \delta^\text{\text{CS}}_t = \left\{
    \begin{matrix*}[l]
    1, \;\;\;\;\;\;\; \mu_t > T_c \\ 
    0, \;\;\;\;\;\;\; \mu_t \leq T_c 
    \end{matrix*}\right., \;\;\;\; t\geq0
    \end{equation}
    \item \textit{Soft-Confidence Switch (SCS)}: Performs a \textit{Confidence Switch} with a smooth transition through a weighted average of the models. By moving farther from the source, i.e. lower confidence, $h_\text{dynamic}$ is weighted more, while, when coming back to the source, increases $h_\text{static}$ contribution. We define two thresholds $T_{\text{s}}$, $T_{\text{d}}$ with $T_{\text{s}} > T_{\text{d}}$ which indicate the $\mu_t$ values where $h_{\text{static}}$ and $h_{\text{dynamic}}$ will be solely used respectively, and we linearly interpolate between the two models when $\mu_t$ is in-between the two thresholds. That is:
    \begin{equation}
        \delta^{\text{SCS}}_t = \max\{\min\{ \frac{1}{T_{\text{s}} - T_{\text{d}}} \mu_t - \frac{T_{\text{d}}}{T_{\text{s}} - T_{\text{d}}}, 1\}, 0\}
    \end{equation}
    \item \textit{Confidence Derivative Switch (CDS)}: Uses the indicator function previously described in Section \ref{sec:3_indicator} to understand if the new domain is farther or closer from the source and selects $h_\text{dynamic}$ or $h_\text{static}$ accordingly.
    \begin{equation}
    \delta^\text{\text{CDS}}_t = \left\{
\begin{matrix*}[l]
1, \;\;\;\;\;\;\; I_t > 0 \\ 
0, \;\;\;\;\;\;\; I_t < 0 \\ 
\delta^\text{CDS}_{t-1}, \;\; I_t=0
\end{matrix*}\right., \;\;\;\; t>0,\;\; \delta_0^\text{CDS}=1
\end{equation}
    \item \textit{Hybrid Switch (HS)}: Sets two thresholds $T_{c_A}$, $T_{c_B}$ with $T_{c_A}>T_{c_B}$ and acts based on confidence values $\mu_t$
    \begin{equation}
        \delta^{\text{HS}}_t = \left\{
    \begin{matrix*}[l]
    1, \;\;\;\;\;\;\;\;\;\;\;\;\;\;\;\;\;\; \mu_t > T_{c_A} \\ 
    \delta_t^\text{CDS}, \;\;T_{c_B} \leq \mu_t \leq T_{c_A} \\ 
    0, \;\;\;\;\;\;\;\;\;\;\;\;\;\;\;\;\;\; \mu_t < T_{c_B}
    \end{matrix*}\right., \;\;\;\; t\geq0
    \end{equation}
The \textit{Hybrid Switch} therefore combines \textit{Confidence Switch} and \textit{Confidence Derivative Switch}: it follows the former for high/low $\mu_t$, the latter otherwise.
\end{itemize}

\section{Experimental Results}
\label{sec:4_result}

The experiments are carried out on the Cityscapes \cite{cordts2016cityscapes} dataset by generating realistic synthetic rain \cite{tremblay2020rain}. In particular, we generate a new training set ($2975$ samples) and validation set ($500$ samples) for each rain intensity. Given a pre-trained model on the original dataset, the online adaptation process takes place by training (without labels) on the rain intensities sequentially. After each pass, the model is validated on all rain intensity validation sets. The experiments include severe rain conditions and show how gradual adaptation compares to direct  -- offline -- adaptation.

\subsection{Baseline Scenario: Increasing Storm}\label{sec:6_base_testing}
As baseline scenario, we use rain intensities of $25$, $50$, $75$, $100$ and $200$. Adaptation happens gradually, from low to high intensities, and then backward until \textit{clear} weather domain, $D_S$, is reached again. We will refer to this adaptation sequence, where we move from source to a sequence of targets and eventually return to the source, as an \textit{adaptation cycle}. 
Each domain counts about 9K frames -- or 5min at 30fps. 
Harder scenarios will be studied in the remainder.

\textbf{Experiment Parameters.}
We use DeepLabv2 \cite{deeplabv2}, which is a common baseline when dealing with domain adaptation on semantic segmentation. The network is although modified to use the ResNet50 \cite{DBLP:journals/corr/HeZRS15} feature extractor instead of the DeepLabv2's default ResNet101 to make training and inference faster. The parameters $\alpha$ and $\beta$ of the SCE are set to $0.1$ and $1$, respectively, while the regularizer parameter $\gamma$ is set to $0.1$.
To measure the accuracy of any model, we compute the mIoU metric. Moreover, on the right most column of each table we report the \textit{harmonic mean} of the overall adaptation process to ease comparison.
Our source code is available at \url{https://github.com/theo2021/OnDA}.

\begin{figure}[t]
    \centering
    \begin{subfigure}[h]{0.32\textwidth}
    \centering
    \scriptsize{(a)} \\
    \includegraphics[width=\textwidth]{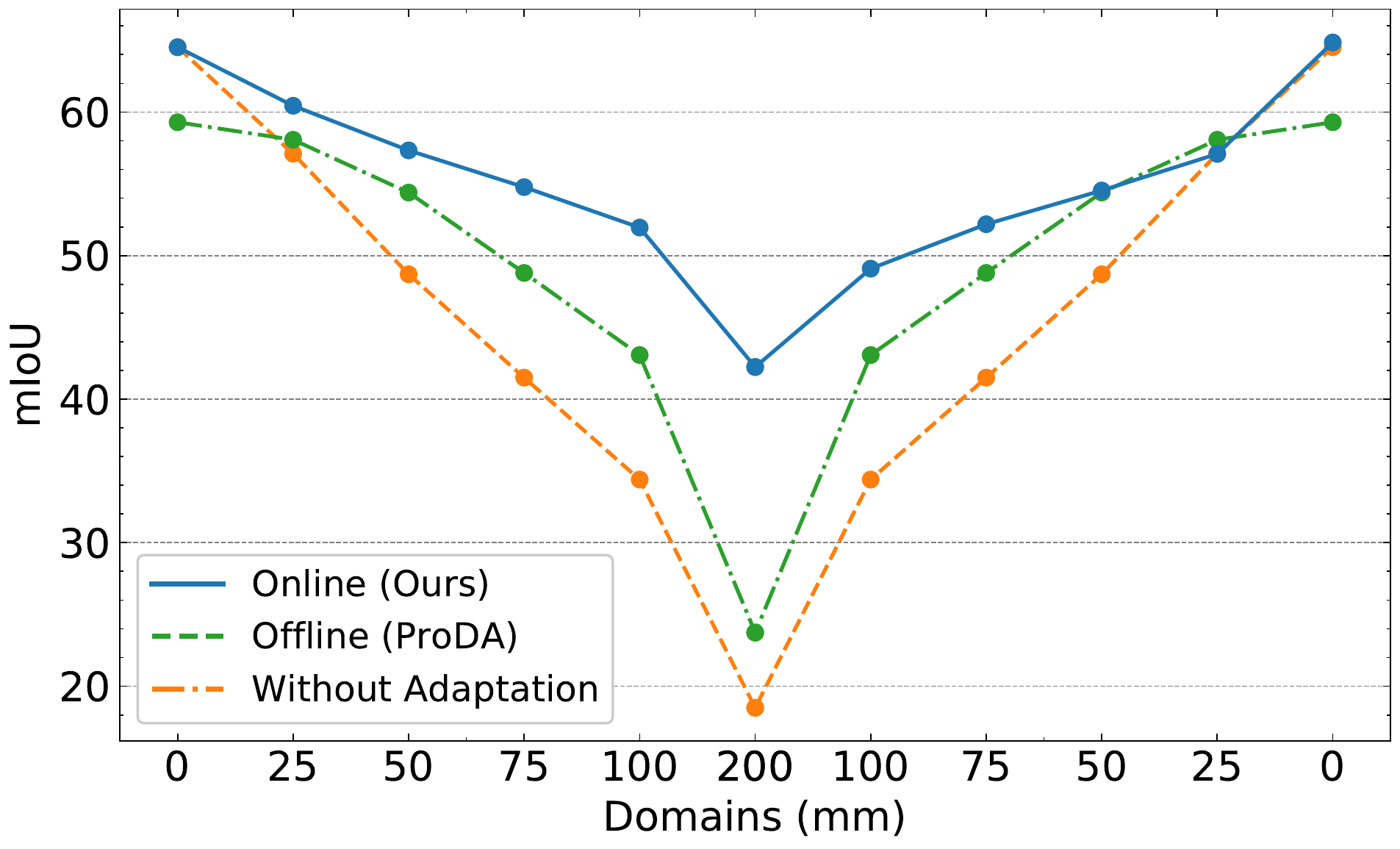}
    \end{subfigure}
    \begin{subfigure}[h]{0.65\textwidth}
    \centering
    \begin{minipage}{0.48\textwidth}
        \centering
            \scriptsize{(b)}
        \end{minipage}
        \begin{minipage}{0.48\textwidth}
        \centering
            \scriptsize{(c)}
        \end{minipage}
    \includegraphics[width=\textwidth]{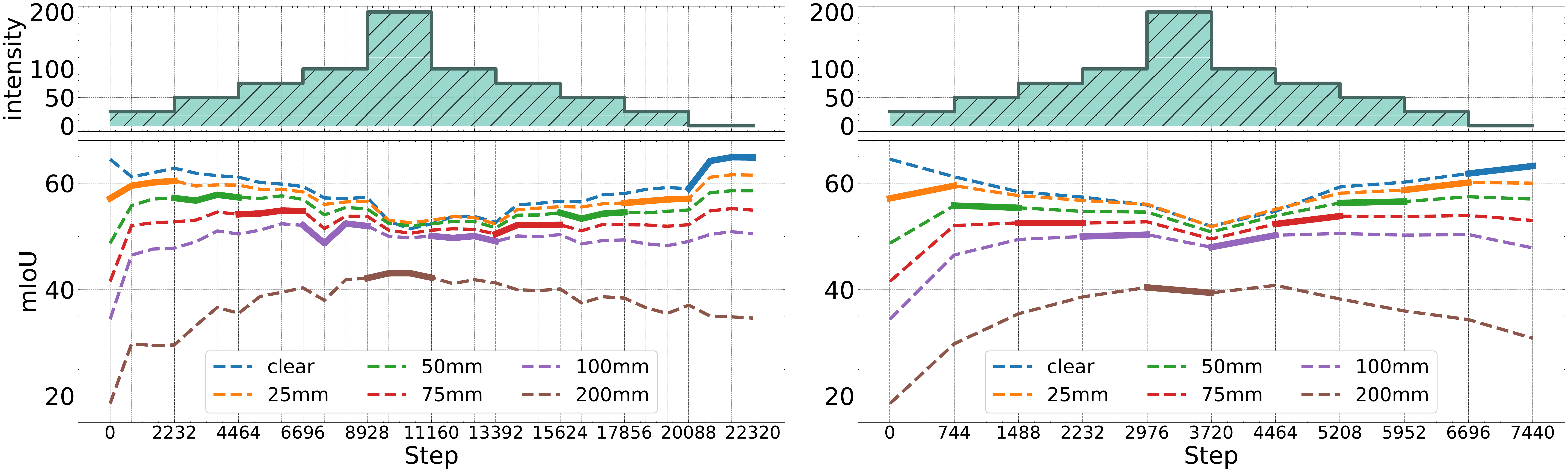}\\
    \end{subfigure}
    \caption{\textbf{Performance comparison and learning process on Increasing Storm.} (a) We plot the mIoU achieved by OnDA using Hybrid Switch (blue), the offline adaptation (green) and the source model (orange), trained on \textit{clear} weather. The offline model is trained using the source domain, and then adapted to all the rainy domains shown in the $x$ axis at once.
    In (b), (c) we show for OnDA, at any given time, mIoU of the model in the currently deployed domain with bold segments. The dashed lines show mIoU over past or future domains.
    }
    \label{fig:6_performance_showcase}
\end{figure}

\subsection{Results on Increasing Storm}
Figure \ref{fig:6_performance_showcase}, on the left, resumes a direct comparison on the Increasing Storm scenario, between the Source model and those adapted either Offline or Online (with the Hybrid Switch). We can notice a higher mIoU achieved by our framework on any domain.
Table \ref{tab:6_main_results} showcases more in detail all the major experiments performed, comparing Test-Time/Online Adaptation, Supervised and Offline Adaptation models in the Increasing Storm scenario. In the Offline experiments, we employ a modified version of \cite{zhang_prototypical_2021} (Stage 1) which is obtained by adopting the Mahalanobis distance (Eq. \ref{eq:3_proto_softmax_pred}) and active BN statistics selection \ref{sec:3_prototypical_pseudolable}, as these improvements result beneficial also in the traditional offline UDA settings. Training is performed until convergence (10 epochs) with decaying learning rate in a standard setup for both Offline and Supervised models, i.e. they have access -- in advance -- to the complete data, hence prototypes can be initialized using the target samples.
The key comparisons are between adaptations methods, either online or offline and how adaptation compares to the fully supervised, ideal case (oracle).
In sum, progressive adaptation sees a significant performance gain compared to directly adapting to a domain offline, as evident by comparing best values (in bold) achieved on each domain. We will now discuss in detail the behavior of the different methods involved in our experiments.

\begin{table}[tt]
\centering
\resizebox{0.99\textwidth}{!}{
\begin{tabular}{ccc}
\begin{tabular}{cclccccccc}
\multicolumn{10}{c}{\textbf{(a)}} \\
\toprule
& &Domain: & clear & 25mm & 50mm & 75mm & 100mm & 200mm & h-mean \\
\midrule
& &Source Model & 64.5 & 57.1 & 48.7 & 41.5 & 34.4 & 18.5 & 37.3 \\
\hline
\hline
\parbox[t]{2mm}{\multirow{8}{*}{\rotatebox[origin=c]{90}{Online}}} & (A) & BN adaptation          &  64.5 & 58.2 & 51.1 & 44.8 &  39.7 &  27.9 & 44.3 \\
& (B) & TENT \cite{wang2021tent} & 64.5 & 57.1 & 48.1 & 41.3 & 33.6 & 15.8 & 35.1 \\
& (C) & TENT + Replay Buffer & 64.5 & 57.6 & 50.0 & 43.8 & 37.4 & 20.5 & 39.7 \\
& (D) & Online Advent                 &  64.5 & 58.7 & 53.5 & 47.6 &  43.0 &  31.1 &   47.0 \\
& (E) & OnDA - Static Model           &  64.5 & \textbf{60.4} & 57.5 & 53.5 &  48.2 &  37.8 &   52.0 \\
& (F) & OnDA - Dynamic Model          &  64.5 & \textbf{60.4} & \textbf{57.8} & 54.7 &  \textbf{52.7} &  41.2 &   54.1 \\
& (G) & OnDA - Confidence Switch      &  64.5 & \textbf{60.4} & 57.5 & 55.1 &  51.3 &  42.1 &   54.1 \\
& (H) & OnDA - Confidence Derivative Switch        &  64.5 & \textbf{60.4} & 57.1 & 54.3 &  52.0 &  \textbf{42.4} &   54.2 \\
& (I) & OnDA - Soft-Confidence Switch &  64.5 & \textbf{60.4} & 57.4 & 54.7 & 52.1 &  42.3 &   \textbf{54.3} \\
& (J) & OnDA - Hybrid Switch          &  64.5 & \textbf{60.4} & 57.3 & \textbf{54.8} &  52.0 &  42.2 &   54.2 \\
& (K) & OnDA - Hybrid Switch One Pass &  64.5 & 59.5 & 55.3 & 52.5 & 50.3 &  39.3 &   52.2
\\
\bottomrule
\end{tabular}
& \quad\quad\quad &

\begin{tabular}{clcccccc}
\multicolumn{8}{c}{\textbf{(b)}} \\
\toprule
& Domain: & 100mm & 75mm & 50mm & 25mm & clear & h-mean \\
\midrule
& Source Model & 34.4 & 41.5 & 48.7 & 57.1 & 64.5 & 37.3 \\
\hline
\hline
(A) & BN adaptation                       &   39.5 & 45.1 & 51.2 & 58.1 & 64.4 & 50.1 \\
(B) & TENT \cite{wang2021tent} & 28.5 & 35.7 & 43.6 & 52.7 & 60.5 & 41.1 \\
(C) & TENT + Replay Buffer & 37.3 & 44.1 & 50.3 & 57.7 & 64.3 & 48.9\\
(D) & Online Advent                       &   43.3 & 48.5 & 54.2 & 58.9 & 64.3 & 52.8 \\
(E) & OnDA - Static Model                        &   47.1 & 50.5 & 52.3 & 56.4 & \textbf{64.8} & 53.6 \\
(F) & OnDA - Dynamic Model                       &   49.8 & 50.1 & 49.9 & 50.3 & 53.3 & 50.6 \\
(G) & OnDA - Confidence Switch                   &   48.3 & 48.8 & 52.7 & 56.0 & 64.6 & 53.5 \\
(H) & OnDA - Confidence Derivative Switch        &   50.1 & 52.5 & 54.4 & 56.6 & 64.7 & 55.2 \\
(I) & OnDA - Soft-Confidence Switch              &   49.3 & 49.7 & 50.1 & 51.8 & 64.2 & 52.5 \\ 
(J) & OnDA - Hybrid Switch                       &   49.1 & 52.2 & 54.5 & 57.1 & \textbf{64.8} & 55.1 \\
(K) & OnDA - Hybrid Switch One Pass              &   \textbf{50.2} & \textbf{53.8} & \textbf{56.5} & \textbf{60.1} & 63.2 & \textbf{56.3} \\
\bottomrule
\end{tabular}
\\
\\
\newlength{\namelength}
\newlength{\numbering}
\setlength{\namelength}{\widthof{Confidence Derivative Switch}}
\setlength{\numbering}{\widthof{a}}
\begin{tabular}{cL{\numbering}L{\namelength}ccccccc}
\multicolumn{10}{c}{\textbf{(c)}} \\
\toprule
& &Domain: & clear & 25mm & 50mm & 75mm & 100mm & 200mm & h-mean \\
\midrule
\parbox[t]{2mm}{\multirow{6}{*}{\rotatebox[origin=c]{90}{Offline}}} && Offline 25mm     &  \textbf{62.8} &   \textbf{60.2} &   \textbf{56.6} &   51.1 &    45.7 &    26.4 &    46.3 \\
& & Offline 50mm         &  60.9 &   59.0 &   55.9 &   \textbf{51.3} &    \textbf{46.4} &    28.9 &    \textbf{47.3} \\
& & Offline 75mm         &  58.8 &   57.2 &   53.6 &   48.5 &    43.8 &    27.2 &    45.0 \\
& & Offline 100mm        &  55.9 &   54.6 &   51.1 &   46.2 &    41.8 &    26.7 &    43.2 \\
& & Offline 200mm        &  49.2 &   50.7 &   49.7 &   47.6 &    45.0 &    \textbf{35.9} &    45.7 \\
& & Offline All          &  59.3 &   58.1 &   54.4 &   48.8 &    43.1 &    23.7 &    43.4 \\
& & Offline All - Advent &  50.8 &   51.7 &   49.1 &   45.9 &    41.9 &    30.9 &    43.6 \\
\bottomrule
\end{tabular} & \quad\quad\quad &

\setlength{\namelength}{\widthof{Confidence Derivative Switch}}
\begin{tabular}{ccL{\namelength}ccccccc}
\multicolumn{10}{c}{\textbf{(d)}} \\
\toprule
& &Domain: & clear & 25mm & 50mm & 75mm & 100mm & 200mm & h-mean \\
\midrule
\parbox[t]{2mm}{\multirow{6}{*}{\rotatebox[origin=c]{90}{Supervised}}} & & Supervised 25mm  &  63.0 &   62.4 &   61.1 &   58.3 &    56.7 &    44.1 &     56.7 \\
& & Supervised 50mm  &  60.4 &   60.6 &   60.4 &   58.2 &    56.9 &    47.4 &     56.9 \\
& & Supervised 75mm  &  56.7 &   58.8 &   58.6 &   57.1 &    56.0 &    48.4 &     55.7 \\
& & Supervised 100mm &  56.5 &   59.0 &   59.9 &   58.3 &    58.0 &    51.8 &     57.1 \\
& & Supervised 200mm &  48.9 &   52.6 &   54.3 &   54.5 &    54.5 &    51.3 &     52.6 \\
& & Supervised All   &  \textbf{64.5}  & \textbf{64.1}  & \textbf{63.7} &  \textbf{63.0} &  \textbf{62.4} &   \textbf{58.2} &   \textbf{62.6} \\
\bottomrule
\end{tabular} \\
\end{tabular}
}

\caption{\textbf{Domain adaptation main results.} Online forward (a), backward (b), Offline (c) and Supervised (d) models are compared. Adaptation happens gradually from low ($25$mm) to high ($200$mm) intensities (a) and backward (b).}
\label{tab:6_main_results}
\end{table}

\textbf{Online Adaptation}. The simplest way to perform adaptation, is by adjusting the BatchNorm statistics (A) in an online manner. Although simple, it manages to yields significant improvements over the source model. Adaptation using TENT \cite{wang2021tent} (B) results less effective compared to simply updating BatchNorm statistics, both in forward (a) and backward (b) adaptation. Introducing the Replay Buffer (C) partially improves the results, yet not surpassing (A).
Online Advent model (D), which is obtained through the use of the Replay Buffer, increases performance even further, yet resulting less performant than self-training approaches.
The Static Model (obtained by fixing $\delta_t=1$) (E) is capable of adapting and reverting to the original domain. Nevertheless, we notice that performance in adaptation can be further increased using a dynamic prior ($\delta_t=0$) (F) -- introduced in Sec. \ref{sec:3_prior_switching}.
Compared to the Static Model (E), the Dynamic Model better adapts to the most challenging domains ($100$ and $200$mm), motivating the need for updating the prior during adaptation.
However, the Dynamic Model is more prone to forgetting: Table \ref{tab:6_main_results} (b), shows performance while gradually returning to the source domain, i.e. retracing domains in reverse order. The Dynamic Model (F) achieves the worst performance once returned to source domain (\textit{clear}).
This issue is solved by switching between the two priors (G-J). 
Among the switching policies, the Confidence Derivative (H) and Hybrid (J) perform the best, increasing adaptation performance substantially ($\sim 24\%$mIoU on the hardest domain). Furthermore, all policies managed to regain the initial performance before adaptation -- see Table \ref{tab:6_main_results} (b). Finally (K) presents the adaptation capabilities of the Hybrid Switch over a 3 times faster Increasing Storm (i.e., happening within fewer frames, as shown in Fig. \ref{fig:6_performance_showcase} (c)): while the forward adaptation achieves marginally lower metrics compared to (J), backward adaptation results more effective on average.

\textbf{Online vs Offline Adaptation.} From Table \ref{tab:6_main_results} (c), it is evident that Offline methods fall short against Online ones (a), proving that it is harder to adapt to the most challenging domains without intermediate adaptations. We have some evidence of this among the entries in the table. Indeed, the best Offline model results on $75$ and $100$mm domains are achieved by the model adapted on $50$mm, suggesting that, sometimes, adapting to an easier domain can be even preferable compared to direct adaptation to the hardest one. Figure \ref{fig:6_performance_showcase} (b, c), outlines the mIoU scores while adapting in an Online manner, on the Increasing Storm setting described so far (b) or by shortening each domain to one third of their length (c). At any time, we also plot the performance for past/future domains (dashed lines). This allows to denote that adapting to close domains (e.g. 50mm) already increases performance to the next to come (75mm, as we can notice from the red dashed line on the left of the bold segment), without yet observing it. Furthermore, the experiment on shorter domains yields similar performance demonstrating, the fast adaptation capabilities of the approach.

\textbf{Adaptation vs Supervised learning.} Although adaptation managed to improve performance, a significant gap between adaptation and supervised learning still exists. Not surprisingly, supervised models perform quite well when fully labelled data is provided, but are always constrained to the source domain, while online adaptation methods can adjust models to new domains on-the-fly.

\subsection{Experiments under additional settings}

In this section, we extend our evaluation by considering different rainy sequences, by studying the impact of the Replay Buffer and by generalizing our framework to different domain changes, such as increasing fog.

\begin{figure*}[t]
    \centering
    \includegraphics[width=0.84\textwidth]{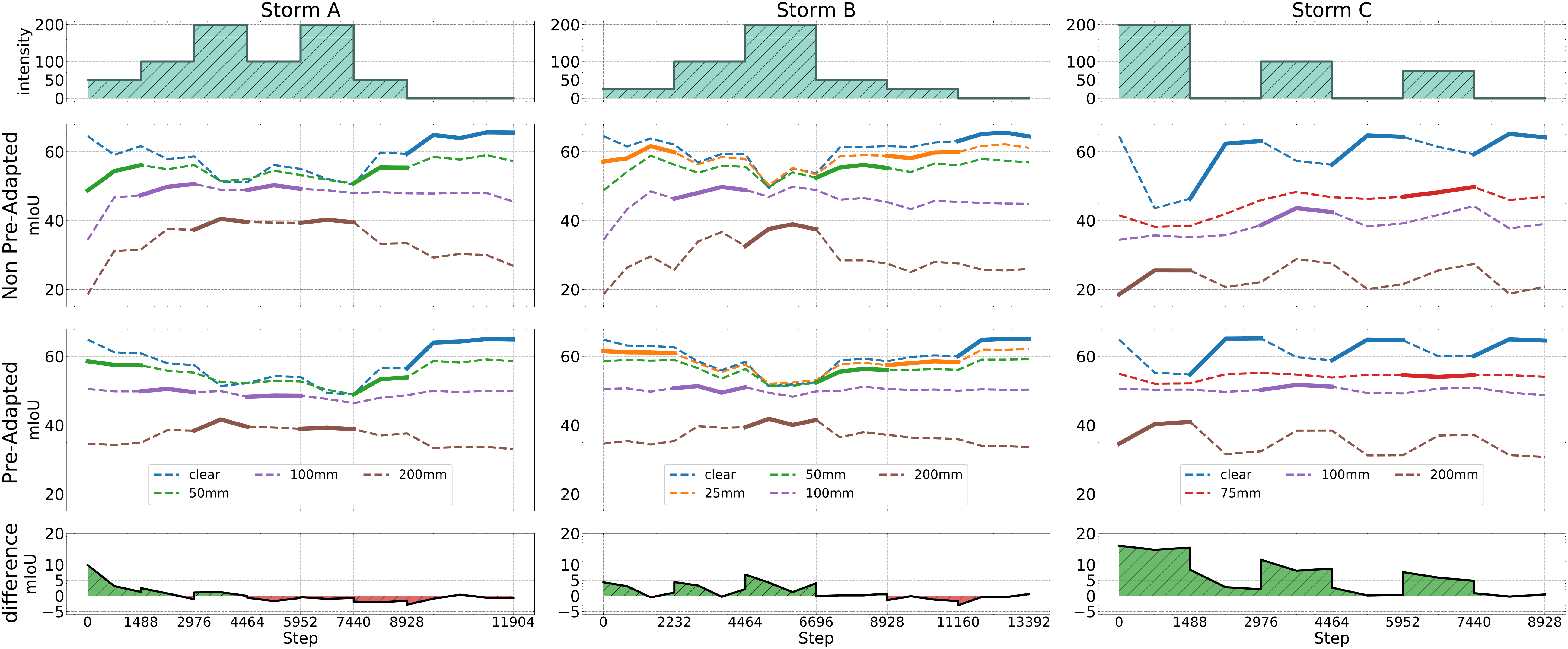}
    \caption{\textbf{Model performance during adaptation}. Experiments on storms A, B \& C. Comparison between starting adaptation from source (Non Pre-Adapted) and after a full \textit{adaptation cycle} on the Increasing Storm (Pre-Adapted).}
    \label{fig:6_A_scenario}
\end{figure*}
\textbf{Evaluation on different Storms.} 
We now run experiments on different rainy scenarios to confirm our previous findings. In particular, we evaluate over three adaptation sequences. In all of them, we use the Hybrid Switch, and we compare two models. The first model is pre-trained on source (\textit{clean} images), while the second model has already experienced a full \textit{adaptation cycle} over the Increasing Storm (Section \ref{sec:6_base_testing}). Results are collected in Figure \ref{fig:6_A_scenario}. On top, we plot histograms describing the three storm intensities, labeled A, B and C and being respectively an \textit{oscillatory} storm (to evaluate OnDA capability of going back and forth in harder domains), a \textit{sudden} storm (with a more aggressive intensity growth) and a \textit{instantaneous} storm (starting with the hardest domain and oscillating significantly).
The plots below show the performance of the two models exposed to the same storm. The last row instead shows the mIoU difference between the two. 
Starting from storm A, we can notice how both models perform similarly and quickly adapt to each domain change.
At bootstrap, the pre-adapted model results better, anyway, the non pre-adapted one quickly catches up, eventually closing the gap between the two.
The same trend occurs on storm B, although the pre-adapted model results more effective during the whole ``forward'' pass.
By looking at storm C, instead, we witness an interesting behavior. Storm C is by far the most challenging in our benchmark due to its abrupt first intensity. The non pre-adapted model fails to adapt to the $200$mm domain encountered at the very beginning, hinting once again that gradual adaptation is preferable. Indeed, the pre-adapted model can instead easily reach the same performance achieved during the Increasing Storm on just a single pass. This result proves that after an \textit{adaptation cycle}, the model is not only able to reach again the source domain with no catastrophic forgetting, but, crucially, it also maintains a memory of the previously experienced domains. Hence, it acquires the ability to cope with more challenging and sudden domain shifts.
Finally, our supplementary material contains qualitative results and refers to a video, showing OnDA in action on the Increasing Storm scenario.

\label{sec:ablation}
\begin{table}[tt]
\centering
\resizebox{0.452\textwidth}{!} 
{\begin{tabular}[t]{cl|rrrrrrrrrrrr}
\multicolumn{14}{c}{(a)} \\
\toprule
& & \multicolumn{2}{c}{clear} & \multicolumn{2}{c}{25mm} & \multicolumn{2}{c}{50mm} & \multicolumn{2}{c}{75mm} & \multicolumn{2}{c}{100mm} & \multicolumn{2}{c}{200mm} \\
\multicolumn{2}{c|}{Buffer} &     \multicolumn{1}{c}{F} &    \multicolumn{1}{c}{B} &    \multicolumn{1}{c}{F} &    \multicolumn{1}{c}{B} &    \multicolumn{1}{c}{F} &    \multicolumn{1}{c}{B} &    \multicolumn{1}{c}{F} &    \multicolumn{1}{c}{B} &     \multicolumn{1}{c}{F} &    \multicolumn{1}{c}{B} &     \multicolumn{1}{c}{F} &    \multicolumn{1}{c}{B} \\
\midrule
(A) & 0     &  64.5 & 57.5 & 60.5 & 55.0 & 57.6 & 53.5 & 54.0 & 50.9 &  50.1 & 49.1 &  41.0 & - \\
(B) & 100   &  64.5 & 63.0 & 60.3 & 56.9 & 56.2 & 54.1 & 54.2 & 51.6 &  51.5 & \textbf{49.3} &  \textbf{42.9} & - \\
(C) & 1000 &  64.5 & 64.8 & 60.4 & \textbf{57.1} & 57.3 & \textbf{54.5} & \textbf{54.8} & \textbf{52.2} &  \textbf{52.0} & 49.1 &  42.2 & - \\
(D) & All   &  64.5 & \textbf{65.4} & \textbf{61.0} & 55.9 & \textbf{58.1} & 54.4 & 53.5 & 51.1 &  51.8 & 49.2 &  41.4 & - \\
\bottomrule
\end{tabular}
}
\resizebox{0.54\textwidth}{!}{\begin{tabular}[t]{cl|rrrrrrrrrrrr}
%& & \multicolumn{4}{c}{Visibility (m)} \\
\multicolumn{14}{c}{(b)} \\
\toprule
& Domain (visibility): & \multicolumn{2}{c}{clear} & \multicolumn{2}{c}{750m} & \multicolumn{2}{c}{375m} & \multicolumn{2}{c}{150m} & \multicolumn{2}{c}{75m} & \multicolumn{2}{c}{h-mean} \\
& {} &     \multicolumn{1}{c}{F} &    \multicolumn{1}{c}{B} &    \multicolumn{1}{c}{F} &    \multicolumn{1}{c}{B} &    \multicolumn{1}{c}{F} &    \multicolumn{1}{c}{B} &    \multicolumn{1}{c}{F} &    \multicolumn{1}{c}{B} &     \multicolumn{1}{c}{F} &    \multicolumn{1}{c}{B} &     \multicolumn{1}{c}{F} &    \multicolumn{1}{c}{B} \\
\midrule
 & Source & \textbf{64.9}& - & 60.9 & - & 54.7 & - & 39.8 & - & 25.2 & - & 43.5& - \\
 & Offline All & 62.4 & - & 62.3 & - & 59.6 & - & 46.8 & - & 31.9 & - & 49.2& -\\
 & OnDA -  Hybrid Switch & \textbf{64.9} & 65.8 & \textbf{63.3} & 62.3 & \textbf{60.7} & 58.8 & \textbf{51.6} & 49.1 & \textbf{42.1} & 42.1 & \textbf{55.1} & 54.1\\
\bottomrule
\end{tabular}}

\caption{\textbf{Additional experiments.} (a): impact of the Replay Buffer on the Increasing Storm cycle using the Hybrid Switch. (b): comparison between Offline and Online adaptation on foggy domains.  F: adaptation from \textit{clear} to the hardest domain, B: backward adaptation (from the hardest domain back to \textit{clear}).}
 \label{tab:6_additional_experiments}
\end{table}

\textbf{Ablation Study - Replay Buffer size.} We now study the impact of the Replay Buffer size, so far set to $1000$ source samples. 
Table \ref{tab:6_additional_experiments} (a) shows a comparison between different Replay Buffer sizes. The model manages to adapt to hard domains even in the absence of a Replay Buffer (A), this although results to a considerable drop in accuracy in the backward phase (about 10\%). With a buffer of $100$ (B) or $1000$ images (C) catastrophic forgetting is solved, while (C) allows for going back to the source with even increased performance. Keeping the whole dataset in the buffer (D) further increases accuracy once back to source, yet not improving adaptation.

\textbf{Additional Case Study - Fog.}
Finally, we test the proposed framework on artificially generated fog \cite{tremblay2020rain} on the Cityscapes training set. The dataset is randomly split into 2475 training and 500 validation samples and we adopted the same experimental set-up presented in the rain scenario. Table \ref{tab:6_additional_experiments} (b) shows a comparison between Source, Offline All and OnDA. Again, our model achieves +10\% mIoU on the hardest domain compared to the one adapted offline, confirming that OnDA can be successfully applied to various domain changes.

\textbf{Limitations.}
Online training requires significant computational resources, which heavily hinder deployment in real-time applications.
We believe that lighter backbones \cite{yu2018bisenet}, efficient training paradigms \cite{tonioni2019real} or selective adaptation can improve this aspect.
From an experimental standpoint, we analyse domain shifts which only affects the input distribution. A larger body of test scenarios, with real data and additional gradual domain shifts would be the ideal stage to assess the performance of OnDA frameworks. 

\section{Summary \& Conclusion}

\label{sec:8_conclusions}

In this paper, we have presented a novel framework for Online Domain Adaptation (OnDA). While state-of-the-art offline adaptation and continuous adaptation methods can successfully tackle limited domain shift, they fall short on cases where there is a significant gap between the source and the deployment domain.
In contrast, we have empirically shown that casting adaptation as an online task and gradually adapting to evolving domains are beneficial for reaching high accuracy on distant domains. We exhaustively evaluated our framework on simulated weather conditions with increasing intensity and in four different kinds of storms, highlighting the robustness of our method in comparison to offline techniques. We believe that our framework will pave the way towards tackling UDA in online manner in the real world.

\textbf{Acknowledgement.} The authors thank Hossein Azizpour, Hedvig Kjellstr{\"o}m and Raoul de Charette for the helpful discussions and guidance.

\bibliographystyle{splncs04}
\bibliography{egbib,morebibs}

\begin{thebibliography}{10}
\providecommand{\url}[1]{\texttt{#1}}
\providecommand{\urlprefix}{URL }
\providecommand{\doi}[1]{https://doi.org/#1}

\bibitem{bengio_curriculum}
Bengio, Y., Louradour, J., Collobert, R., Weston, J.: Curriculum learning. In:
  Proceedings of the 26th Annual International Conference on Machine Learning.
  p. 41–48. ICML '09, Association for Computing Machinery, New York, NY, USA
  (2009). \doi{10.1145/1553374.1553380},
  \url{https://doi.org/10.1145/1553374.1553380}

\bibitem{bobu_adapting_2018}
Bobu, A., Hoffman, J., Tzeng, E., Darrell, T.: Adapting to continuously
  shifting domains. In: {ICLR} 2018 Workshop Program Chairs (2018),
  \url{https://openreview.net/forum?id=BJsBjPJvf}, 00000

\bibitem{Cardace_2022_WACV}
Cardace, A., Zama~Ramirez, P., Salti, S., Di~Stefano, L.: Shallow features
  guide unsupervised domain adaptation for semantic segmentation at class
  boundaries. In: Proceedings of the IEEE/CVF Winter Conference on Applications
  of Computer Vision (WACV). pp. 1160--1170 (January 2022)

\bibitem{chang_domain-specific_2019}
Chang, W.G., You, T., Seo, S., Kwak, S., Han, B.: Domain-specific batch
  normalization for unsupervised domain adaptation. CoRR  (2019),
  \url{http://arxiv.org/abs/1906.03950}

\bibitem{chen2019progressive}
Chen, C., Xie, W., Huang, W., Rong, Y., Ding, X., Huang, Y., Xu, T., Huang, J.:
  Progressive feature alignment for unsupervised domain adaptation. In:
  Proceedings of the IEEE/CVF Conference on Computer Vision and Pattern
  Recognition. pp. 627--636 (2019)

\bibitem{chen2020naive}
Chen, L.C., Lopes, R.G., Cheng, B., Collins, M.D., Cubuk, E.D., Zoph, B., Adam,
  H., Shlens, J.: Naive-student: Leveraging semi-supervised learning in video
  sequences for urban scene segmentation. In: European Conference on Computer
  Vision. pp. 695--714. Springer (2020)

\bibitem{chen2017deeplab}
Chen, L.C., Papandreou, G., Kokkinos, I., Murphy, K., Yuille, A.L.: Deeplab:
  Semantic image segmentation with deep convolutional nets, atrous convolution,
  and fully connected crfs. IEEE transactions on pattern analysis and machine
  intelligence  \textbf{40}(4),  834--848 (2017)

\bibitem{deeplabv2}
Chen, L.C., Papandreou, G., Kokkinos, I., Murphy, K., Yuille, A.L.: Deeplab:
  Semantic image segmentation with deep convolutional nets, atrous convolution,
  and fully connected crfs. IEEE Transactions on Pattern Analysis and Machine
  Intelligence  \textbf{40}(4),  834–848 (Apr 2018).
  \doi{10.1109/tpami.2017.2699184},
  \url{http://dx.doi.org/10.1109/TPAMI.2017.2699184}

\bibitem{chen_no_2017}
Chen, Y.H., Chen, W.Y., Chen, Y.T., Tsai, B.C., Wang, Y.C.F., Sun, M.: No more
  discrimination: Cross city adaptation of road scene segmenters. In: 2017
  {IEEE} International Conference on Computer Vision ({ICCV}). pp. 2011--2020.
  {IEEE} (2017). \doi{10.1109/ICCV.2017.220},
  \url{http://ieeexplore.ieee.org/document/8237482/}, 00000

\bibitem{cicek_unsupervised_2019}
Cicek, S., Soatto, S.: Unsupervised domain adaptation via regularized
  conditional alignment. CoRR  (2019), \url{http://arxiv.org/abs/1905.10885},
  00000

\bibitem{cordts2016cityscapes}
Cordts, M., Omran, M., Ramos, S., Rehfeld, T., Enzweiler, M., Benenson, R.,
  Franke, U., Roth, S., Schiele, B.: The cityscapes dataset for semantic urban
  scene understanding (2016)

\bibitem{stylization}
{Dundar}, A., {Liu}, M.Y., {Yu}, Z., {Wang}, T.C., {Zedlewski}, J., {Kautz},
  J.: Domain stylization: A fast covariance matching framework towards domain
  adaptation. IEEE Transactions on Pattern Analysis and Machine Intelligence
  pp.~1--1 (2020). \doi{10.1109/TPAMI.2020.2969421}

\bibitem{furlanello2018born}
Furlanello, T., Lipton, Z., Tschannen, M., Itti, L., Anandkumar, A.: Born again
  neural networks. In: International Conference on Machine Learning. pp.
  1607--1616. PMLR (2018)

\bibitem{ganin}
Ganin, Y., Ustinova, E., Ajakan, H., Germain, P., Larochelle, H., Laviolette,
  F., Marchand, M., Lempitsky, V.: Domain-adversarial training of neural
  networks. The journal of machine learning research  \textbf{17}(1),
  2096–2030 (Jan 2016)

\bibitem{DBLP:journals/corr/HeZRS15}
He, K., Zhang, X., Ren, S., Sun, J.: Deep residual learning for image
  recognition. CoRR  \textbf{abs/1512.03385} (2015),
  \url{http://arxiv.org/abs/1512.03385}

\bibitem{cycada}
Hoffman, J., Tzeng, E., Park, T., Zhu, J.Y., Isola, P., Saenko, K., Efros, A.,
  Darrell, T.: {C}y{CADA}: Cycle-consistent adversarial domain adaptation. In:
  Dy, J., Krause, A. (eds.) Proceedings of the 35th International Conference on
  Machine Learning. Proceedings of Machine Learning Research, vol.~80, pp.
  1989--1998. PMLR, Stockholmsmässan, Stockholm Sweden (10--15 Jul 2018)

\bibitem{hoffman_fcns_2016}
Hoffman, J., Wang, D., Yu, F., Darrell, T.: {FCNs} in the wild: Pixel-level
  adversarial and constraint-based adaptation. CoRR  (2016),
  \url{http://arxiv.org/abs/1612.02649}, 00000

\bibitem{hoyer2021daformer}
Hoyer, L., Dai, D., Van~Gool, L.: Daformer: Improving network architectures and
  training strategies for domain-adaptive semantic segmentation. arXiv preprint
  arXiv:2111.14887  (2021)

\bibitem{huang2017arbitrary}
Huang, X., Belongie, S.: Arbitrary style transfer in real-time with adaptive
  instance normalization. In: Proceedings of the IEEE International Conference
  on Computer Vision. pp. 1501--1510 (2017)

\bibitem{ioffe2015batch}
Ioffe, S., Szegedy, C.: Batch normalization: Accelerating deep network training
  by reducing internal covariate shift. In: International conference on machine
  learning. pp. 448--456. PMLR (2015)

\bibitem{iwasawa2021testtime}
Iwasawa, Y., Matsuo, Y.: Test-time classifier adjustment module for
  model-agnostic domain generalization. In: Beygelzimer, A., Dauphin, Y.,
  Liang, P., Vaughan, J.W. (eds.) Advances in Neural Information Processing
  Systems (2021)

\bibitem{ltir}
Kim, M., Byun, H.: Learning texture invariant representation for domain
  adaptation of semantic segmentation. 2020 IEEE/CVF Conference on Computer
  Vision and Pattern Recognition (CVPR)  (Jun 2020).
  \doi{10.1109/cvpr42600.2020.01299},
  \url{http://dx.doi.org/10.1109/cvpr42600.2020.01299}

\bibitem{klingner2022unsupervised}
Klingner, M., Term{\"o}hlen, J.A., Ritterbach, J., Fingscheidt, T.:
  Unsupervised batchnorm adaptation (ubna): A domain adaptation method for
  semantic segmentation without using source domain representations. In:
  Proceedings of the IEEE/CVF Winter Conference on Applications of Computer
  Vision. pp. 210--220 (2022)

\bibitem{kuznietsov2022towards}
Kuznietsov, Y., Proesmans, M., Van~Gool, L.: Towards unsupervised online domain
  adaptation for semantic segmentation. In: Proceedings of the IEEE/CVF Winter
  Conference on Applications of Computer Vision. pp. 261--271 (2022)

\bibitem{lao2020continuous}
Lao, Q., Jiang, X., Havaei, M., Bengio, Y.: Continuous domain adaptation with
  variational domain-agnostic feature replay. arXiv preprint arXiv:2003.04382
  (2020)

\bibitem{pseudolabel}
Lee, D.: Pseudo-label : The simple and efficient semi-supervised learning
  method for deep neural networks. In: Workshop on challenges in representation
  learning, ICML (2013)

\bibitem{bdl}
Li, Y., Yuan, L., Vasconcelos, N.: Bidirectional learning for domain adaptation
  of semantic segmentation. 2019 IEEE/CVF Conference on Computer Vision and
  Pattern Recognition (CVPR)  (Jun 2019). \doi{10.1109/cvpr.2019.00710},
  \url{http://dx.doi.org/10.1109/CVPR.2019.00710}

\bibitem{liang_we_2020}
Liang, J., Hu, D., Feng, J.: Do we really need to access the source data?
  source hypothesis transfer for unsupervised domain adaptation. CoRR  (2020),
  \url{http://arxiv.org/abs/2002.08546}

\bibitem{liu_source-free_2021}
Liu, Y., Zhang, W., Wang, J.: Source-free domain adaptation for semantic
  segmentation  (2021), \url{http://arxiv.org/abs/2103.16372}

\bibitem{liu2021ttt}
Liu, Y., Kothari, P., van Delft, B.G., Bellot-Gurlet, B., Mordan, T., Alahi,
  A.: {TTT}++: When does self-supervised test-time training fail or thrive? In:
  Beygelzimer, A., Dauphin, Y., Liang, P., Vaughan, J.W. (eds.) Advances in
  Neural Information Processing Systems (2021)

\bibitem{liu_open_2020}
Liu, Z., Miao, Z., Pan, X., Zhan, X., Lin, D., Yu, S.X., Gong, B.: Open
  compound domain adaptation. In: 2020 {IEEE}/{CVF} Conference on Computer
  Vision and Pattern Recognition ({CVPR}). pp. 12403--12412. {IEEE} (2020).
  \doi{10.1109/CVPR42600.2020.01242},
  \url{https://ieeexplore.ieee.org/document/9157145/}

\bibitem{iast}
Mei, K., Zhu, C., Zou, J., Zhang, S.: Instance adaptive self-training for
  unsupervised domain adaptation. Lecture Notes in Computer Science p.
  415–430 (2020)

\bibitem{Pan_2020}
Pan, F., Shin, I., Rameau, F., Lee, S., Kweon, I.S.: Unsupervised intra-domain
  adaptation for semantic segmentation through self-supervision. 2020 IEEE/CVF
  Conference on Computer Vision and Pattern Recognition (CVPR)  (Jun 2020).
  \doi{10.1109/cvpr42600.2020.00382},
  \url{http://dx.doi.org/10.1109/cvpr42600.2020.00382}

\bibitem{poularikas_handbook_1999}
Poularikas, A.D.: The handbook of formulas and tables for signal processing.
  The electrical engineering handbook series, {CRC} Press ; Springer : {IEEE}
  Press (1999)

\bibitem{sakaridis_model_2018}
Sakaridis, C., Dai, D., Hecker, S., Van~Gool, L.: Model adaptation with
  synthetic and real data for semantic dense foggy scene understanding  (2018),
  \url{http://arxiv.org/abs/1808.01265}

\bibitem{sohn_fixmatch_nodate}
Sohn, K., Berthelot, D., Li, C., Zhang, Z., Carlini, N., Cubuk, E.D., Kurakin,
  A., Zhang, H., Raffel, C.: Fixmatch: Simplifying semi-supervised learning
  with consistency and confidence. CoRR  \textbf{abs/2001.07685} (2020),
  \url{https://arxiv.org/abs/2001.07685}

\bibitem{Stan2021UnsupervisedMA}
Stan, S., Rostami, M.: Unsupervised model adaptation for continual semantic
  segmentation. In: AAAI (2021)

\bibitem{su_gradient_2020}
Su, P., Tang, S., Gao, P., Qiu, D., Zhao, N., Wang, X.: Gradient regularized
  contrastive learning for continual domain adaptation  (2020),
  \url{http://arxiv.org/abs/2007.12942}, 00000

\bibitem{shift}
Sun, T., Segu, M., Postels, J., Wang, Y., Van~Gool, L., Schiele, B., Tombari,
  F., Yu, F.: {SHIFT:} a synthetic driving dataset for continuous multi-task
  domain adaptation. In: Computer Vision and Pattern Recognition (2022)

\bibitem{tonioni2019real}
Tonioni, A., Tosi, F., Poggi, M., Mattoccia, S., Stefano, L.D.: Real-time
  self-adaptive deep stereo. In: Proceedings of the IEEE/CVF Conference on
  Computer Vision and Pattern Recognition. pp. 195--204 (2019)

\bibitem{tremblay2020rain}
Tremblay, M., Halder, S.S., de~Charette, R., Lalonde, J.F.: Rain rendering for
  evaluating and improving robustness to bad weather. International Journal of
  Computer Vision  (2020)

\bibitem{adaptsegnet}
Tsai, Y.H., Hung, W.C., Schulter, S., Sohn, K., Yang, M.H., Chandraker, M.:
  Learning to adapt structured output space for semantic segmentation. 2018
  IEEE/CVF Conference on Computer Vision and Pattern Recognition  (Jun 2018).
  \doi{10.1109/cvpr.2018.00780},
  \url{http://dx.doi.org/10.1109/CVPR.2018.00780}

\bibitem{vu2019dada}
Vu, T.H., Jain, H., Bucher, M., Cord, M., P{\'e}rez, P.: Dada: Depth-aware
  domain adaptation in semantic segmentation. In: ICCV (2019)

\bibitem{advent}
Vu, T.H., Jain, H., Bucher, M., Cord, M., Perez, P.: Advent: Adversarial
  entropy minimization for domain adaptation in semantic segmentation. 2019
  IEEE/CVF Conference on Computer Vision and Pattern Recognition (CVPR)  (Jun
  2019). \doi{10.1109/cvpr.2019.00262},
  \url{http://dx.doi.org/10.1109/CVPR.2019.00262}

\bibitem{vu_advent_2019}
Vu, T.H., Jain, H., Bucher, M., Cord, M., Pérez, P.: {ADVENT}: Adversarial
  entropy minimization for domain adaptation in semantic segmentation  (2019),
  \url{http://arxiv.org/abs/1811.12833}, 00000

\bibitem{wang2021tent}
Wang, D., Shelhamer, E., Liu, S., Olshausen, B., Darrell, T.: Tent: Fully
  test-time adaptation by entropy minimization. In: International Conference on
  Learning Representations (2021)

\bibitem{fada}
Wang, H., Shen, T., Zhang, W., Duan, L., Mei, T.: Classes matter: A
  fine-grained adversarial approach to cross-domain semantic segmentation. In:
  The European Conference on Computer Vision (ECCV) (August 2020)

\bibitem{wang_symmetric_2019}
Wang, Y., Ma, X., Chen, Z., Luo, Y., Yi, J., Bailey, J.: Symmetric cross
  entropy for robust learning with noisy labels  (2019),
  \url{http://arxiv.org/abs/1908.06112}

\bibitem{dcan}
Wu, Z., Han, X., Lin, Y.L., Uzunbas, M.G., Goldstein, T., Lim, S.N., Davis,
  L.S.: Dcan: Dual channel-wise alignment networks for unsupervised scene
  adaptation. Lecture Notes in Computer Science p. 535–552 (2018)

\bibitem{wu_ace_2019}
Wu, Z., Wang, X., Gonzalez, J., Goldstein, T., Davis, L.: {ACE}: Adapting to
  changing environments for semantic segmentation. In: 2019 {IEEE}/{CVF}
  International Conference on Computer Vision ({ICCV}). pp. 2121--2130. {IEEE}
  (2019). \doi{10.1109/ICCV.2019.00221},
  \url{https://ieeexplore.ieee.org/document/9009823/}

\bibitem{wulfmeier_incremental_2018}
Wulfmeier, M., Bewley, A., Posner, I.: Incremental adversarial domain
  adaptation for continually changing environments  (2018),
  \url{http://arxiv.org/abs/1712.07436}, 00000

\bibitem{yang_fda_2020}
Yang, Y., Soatto, S.: {FDA}: Fourier domain adaptation for semantic
  segmentation. In: 2020 {IEEE}/{CVF} Conference on Computer Vision and Pattern
  Recognition ({CVPR}). pp. 4084--4094. {IEEE} (2020).
  \doi{10.1109/CVPR42600.2020.00414},
  \url{https://ieeexplore.ieee.org/document/9157228/}

\bibitem{yu2018bisenet}
Yu, C., Wang, J., Peng, C., Gao, C., Yu, G., Sang, N.: Bisenet: Bilateral
  segmentation network for real-time semantic segmentation. In: Proceedings of
  the European conference on computer vision (ECCV). pp. 325--341 (2018)

\bibitem{yuan2021segmentation}
Yuan, Y., Chen, X., Chen, X., Wang, J.: Segmentation transformer:
  Object-contextual representations for semantic segmentation. In: European
  Conference on Computer Vision (ECCV). vol.~1 (2021)

\bibitem{zhai2019s4l}
{Zhai}, X., {Oliver}, A., {Kolesnikov}, A., {Beyer}, L.: {S4L: Self-Supervised
  Semi-Supervised Learning}. arXiv e-prints arXiv:1905.03670 (May 2019)

\bibitem{zhang_prototypical_2021}
Zhang, P., Zhang, B., Zhang, T., Chen, D., Wang, Y., Wen, F.: Prototypical
  pseudo label denoising and target structure learning for domain adaptive
  semantic segmentation  (2021), \url{http://arxiv.org/abs/2101.10979}

\bibitem{zhang_category_2019}
Zhang, Q., Zhang, J., Liu, W., Tao, D.: Category anchor-guided unsupervised
  domain adaptation for semantic segmentation. Advances in Neural Information
  Processing Systems  (2019), \url{http://arxiv.org/abs/1910.13049}

\bibitem{mrnet}
Zheng, Z., Yang, Y.: Unsupervised scene adaptation with memory regularization
  in vivo. Proceedings of the Twenty-Ninth International Joint Conference on
  Artificial Intelligence  (Jul 2020). \doi{10.24963/ijcai.2020/150},
  \url{http://dx.doi.org/10.24963/ijcai.2020/150}

\bibitem{cyclegan}
Zhu, J.Y., Park, T., Isola, P., Efros, A.A.: Unpaired image-to-image
  translation using cycle-consistent adversarial networks. 2017 IEEE
  International Conference on Computer Vision (ICCV)  (Oct 2017).
  \doi{10.1109/iccv.2017.244}, \url{http://dx.doi.org/10.1109/ICCV.2017.244}

\bibitem{zou2018unsupervised}
Zou, Y., Yu, Z., Kumar, B., Wang, J.: Unsupervised domain adaptation for
  semantic segmentation via class-balanced self-training. In: Proceedings of
  the European conference on computer vision (ECCV). pp. 289--305 (2018)

\bibitem{cbst}
Zou, Y., Yu, Z., Liu, X., Kumar, B.V.K.V., Wang, J.: Confidence regularized
  self-training. 2019 IEEE/CVF International Conference on Computer Vision
  (ICCV)  (Oct 2019). \doi{10.1109/iccv.2019.00608},
  \url{http://dx.doi.org/10.1109/ICCV.2019.00608}

\bibitem{zou2019confidence}
Zou, Y., Yu, Z., Liu, X., Kumar, B., Wang, J.: Confidence regularized
  self-training. In: Proceedings of the IEEE/CVF International Conference on
  Computer Vision. pp. 5982--5991 (2019)

\end{thebibliography}

\phantom{Supplementary}
\multido{\i=1+1}{9}{


\section*{Supplementary material}

# Online Domain Adaptation for Semantic Segmentation in Ever-Changing Conditions – Supplementary Material

Theodoros Panagiotakopoulos[1*]     Pier Luigi Dovesi[2]     Linus Härenstam-Nielsen[3,4*]
Matteo Poggi[5]

[1]King     [2]Univrses     [3]Kudan     [4] Technical University of Munich
[5]University of Bologna

This document introduces further details on the ECCV 2022 paper - "Online Domain Adaptation for Semantic Segmentation in Ever-Changing Conditions". In Section 1, we present results of the OnDA framework on the CityScapes dataset [2] with qualitative samples and a video sequence, in Section 2 we detail the model implementation and hyperparameters, while in Section 3 we further describe the working principle behind Domain Shift Detection. Then we provide additional explanation on some experimental results presented in the paper, in particular we focus on the Standard deviation, Section 4, and the Calibration Error, Section 5. Finally, in Section 6 we expand our Ablation Study and study the impact of the learning rate, the effect of the confidence regularization and Batch Normalization.

## 1. CityScapes qualitative comparison between Source, Offline and Online Adapted Model.

We provide additional, qualitative results concerning our experiments on the Increasing Storm scenario. The whole adaptation process is also featured in a video sequence (23.46 minutes).

A shorter version of the video (3 minutes) is also available.

- **short version**: : https://youtu.be/cany-lUNWY8

- **long version**: https://youtu.be/igtmgafiurY

### 1.1. Video Analysis

To reproduce a realistic online adaptation set-up, we employed the Frankfurt CityScapes video sequence. We divided it in equally sized intervals and then we rendered the rain according the Increasing Storm schema *i.e.* 0mm (*clear*), 25mm, 50mm, 75mm, 100mm, 200mm, 100mm, 75mm, 50mm, 25mm, 0mm. We sub-sample the video to 10fps, with a total length of 23.46 minutes, and the duration of each target domain is 3239 frames, corresponding to around 5.4 minutes (the video is accelerated 2.5 times, hence every domain is presented in 2.09 minutes). It is worth mentioning that the adopted rain rendering methodology [6] does not limit to generating rain random particles, as it uses monocular depth prediction [3] and depth refinement [1] to estimate the rain occlusion in the image. Moreover, other effects are employed to make the sky darker and thus emulate the corresponding cloud coverage.

The OnDA framework employs the Hybrid Switch with same configuration presented in all the other experiments. The only differences consisted in a minor update of the two Hybrid Switch thresholds ($T_{c_A}$ and $T_{c_B}$) and in the introduction of a debouncing window on the switching indicator function, $I_t$, as introduced in Section 3.2 of the paper.

In Figure 1 we display video frames and the corresponding segmentation masks collected at regular time intervals so to cover all the deployment domains. The video sequence does not present ground-truth for a quantitative evaluation, anyway the qualitative analysis presents a clear pattern and we feel it significant of a real, online adaptation scenario. The Source model is steadily degraded as the rain intensifies, presenting visible degradation right from the 25mm rain and completely collapsing in the 100mm and 200mm domains. The Offline model instead has been pre-adapted offline on all the rain intensities (for this task we used the CityScapes train split). The Offline model is visibly more robust to the easier targets, while struggling in the 100mm case and collapsing in the 200mm. Finally, the Online model performs similarly or marginally better than the Offline method in all easy targets, while still being robust to the 100mm and 200mm cases.

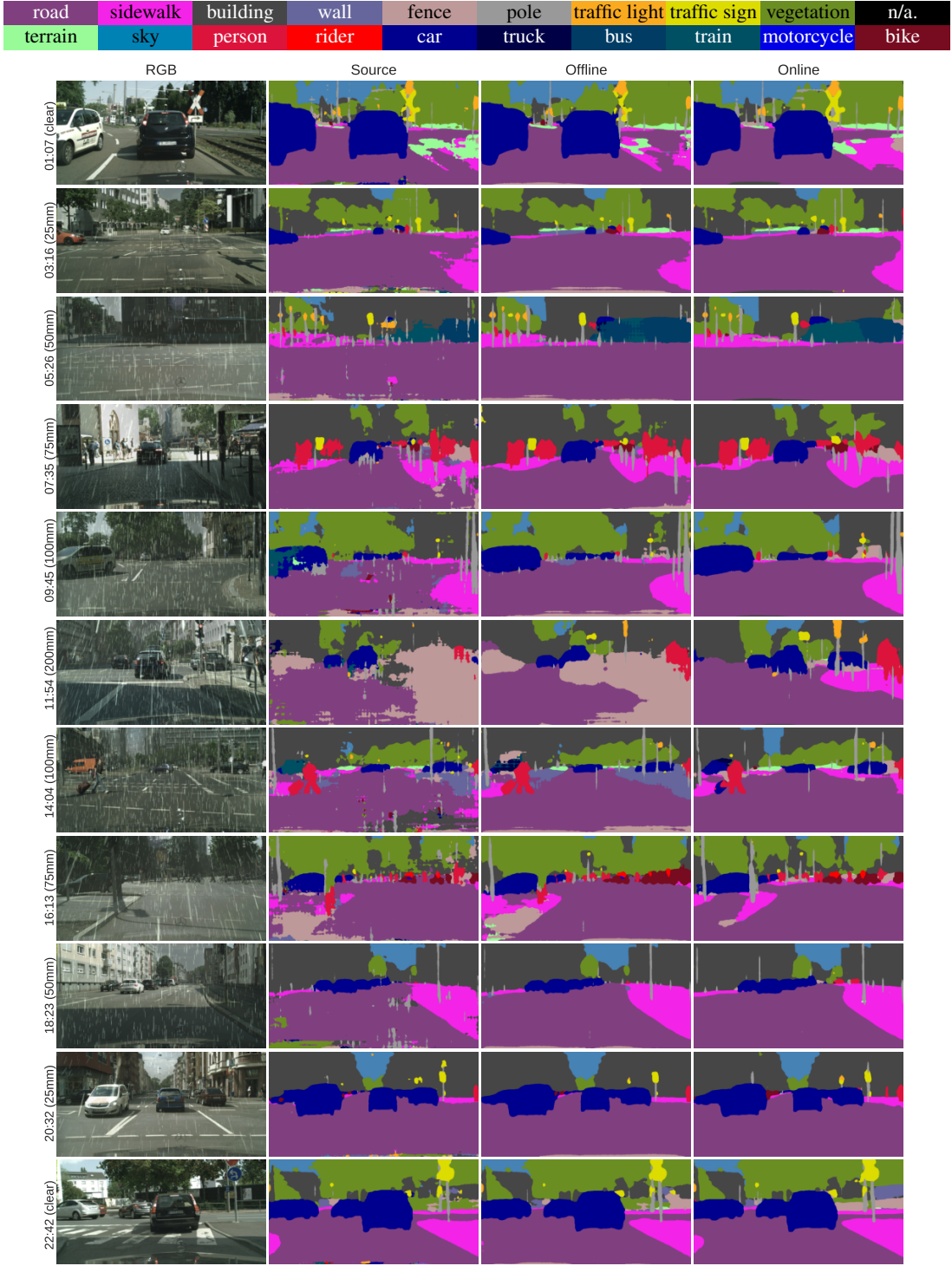

Figure 1. **Samples from the Cityscapes Frankfurt video using artificial rain and comparing the Source, Offline and Online Adapted Model.** The Online model gets adapted on-the-fly exploiting the video footage itself. Images start from frame 1677 (1:07) and are captured every 3239 frames (∼2:09).

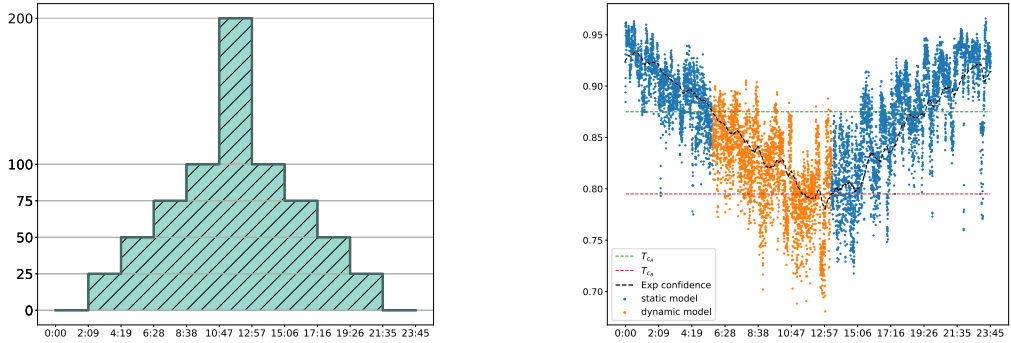

Figure 2. **Domain sequence and switching policy – Increasing Storm.** We show rain intensity over time (left), confidence and switching policy behavior (right) when adapting over the Increasing Storm.

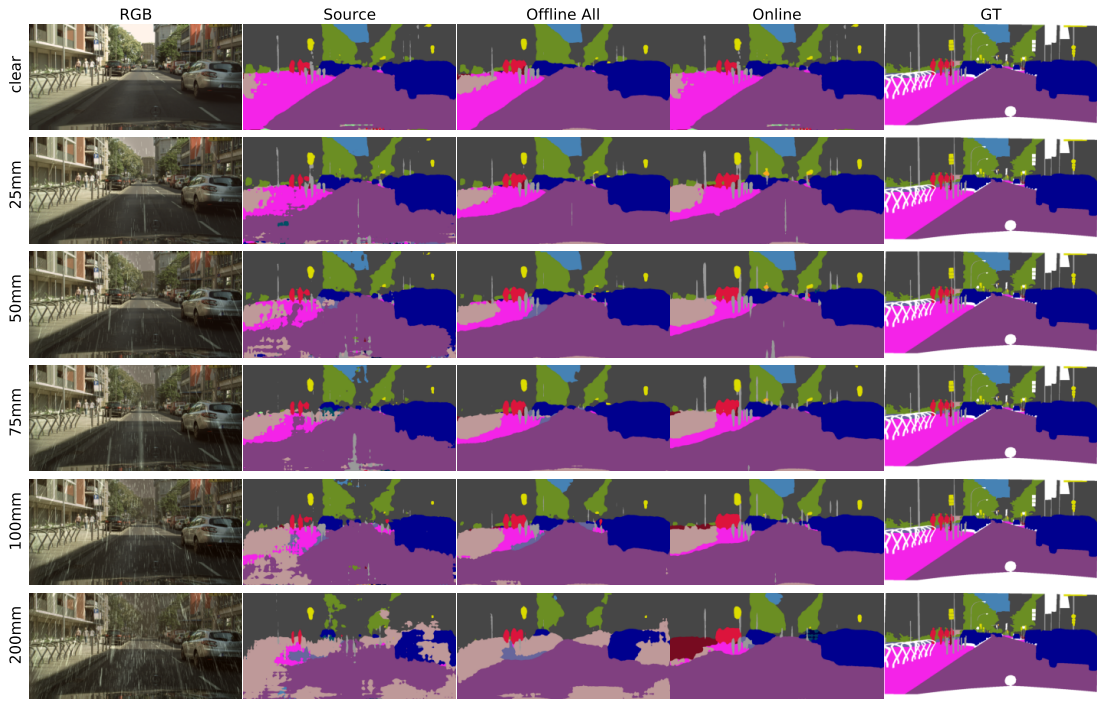

Figure 3. **Qualitative comparison of the Source, Offline and Online Models on rainy images.** From top to bottom, we report a single sample from each domain, from *clear* to 200mm rain intensity.

In Figure 2 we present the rain intensities of the Increasing Storm and the Hybrid Switch mechanism in action over that sequence. In particular we notice how the static prior, $h_{\text{static}}$, is rapidly switched for the dynamic one, $h_{\text{dynamic}}$, during the 50mm domain. This is the consequence of $I_t$ being negative (as $z_t$ is indeed decreasing), since the model is moving further from the source domain. We notice how the model keeps (and updates) $h_{\text{dynamic}}$ until the "backward phase" starts, then as the confidence of $h_{\text{static}}$ grows – leading to a positive $I_t$ – the model prior is switched to $h_{\text{static}}$ again. More details about the switching principle are reported in Sec. 3.

### 1.2. Validation Set Qualitatives

In Figure 3 and 4 we include more qualitative samples taken from the CityScapes validation set. In Figure 5 we present qualitative samples with fog. In any of these figures, the samples present the comparison between Source, Offline All, Online (Hybrid Switch), and Ground-Truth (GT).

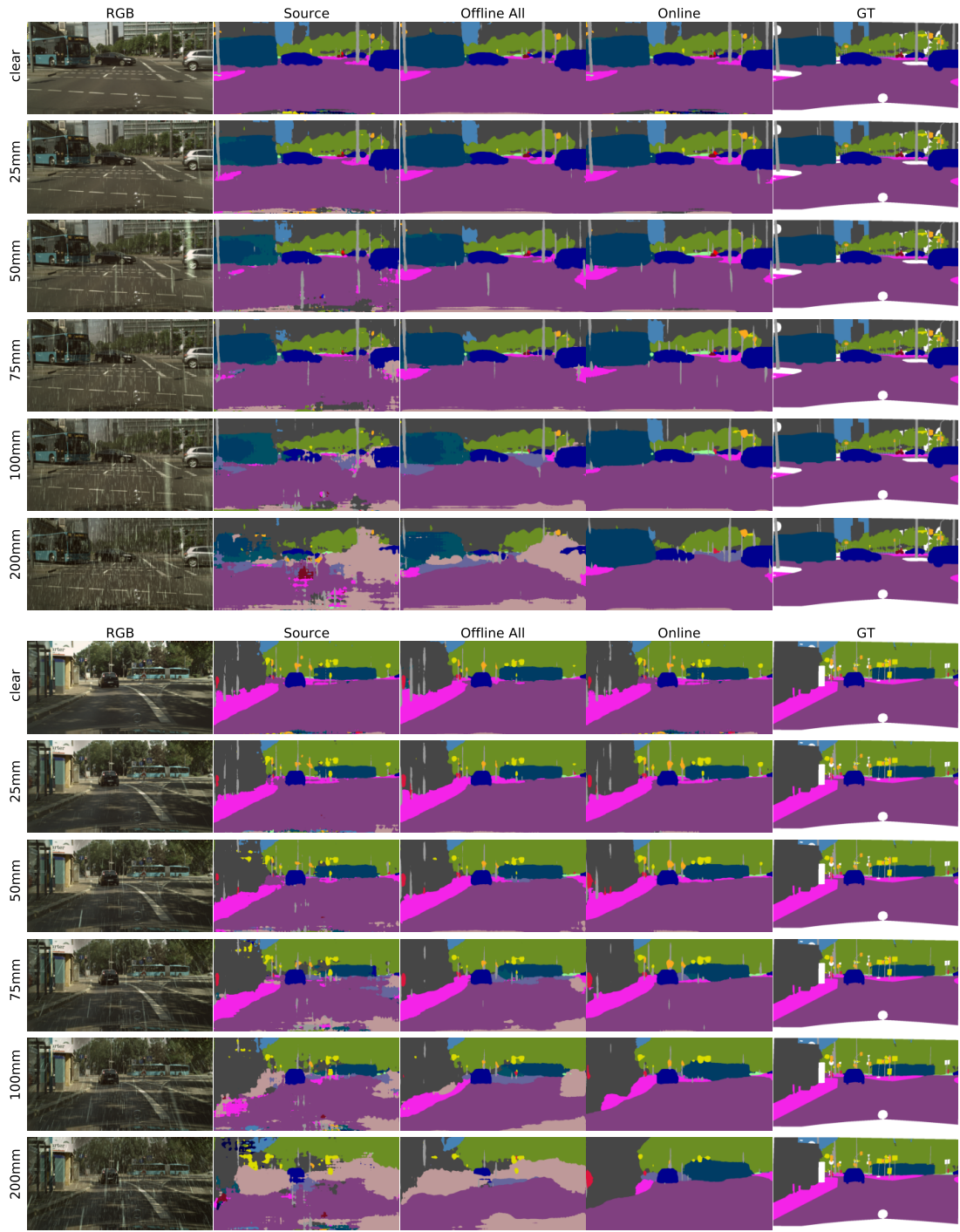

Figure 4. **Qualitative comparison of the Source, Offline and Online Models on rainy images.** From top to bottom, we report a single sample from each domain, from *clear* to 200mm rain intensity.

4

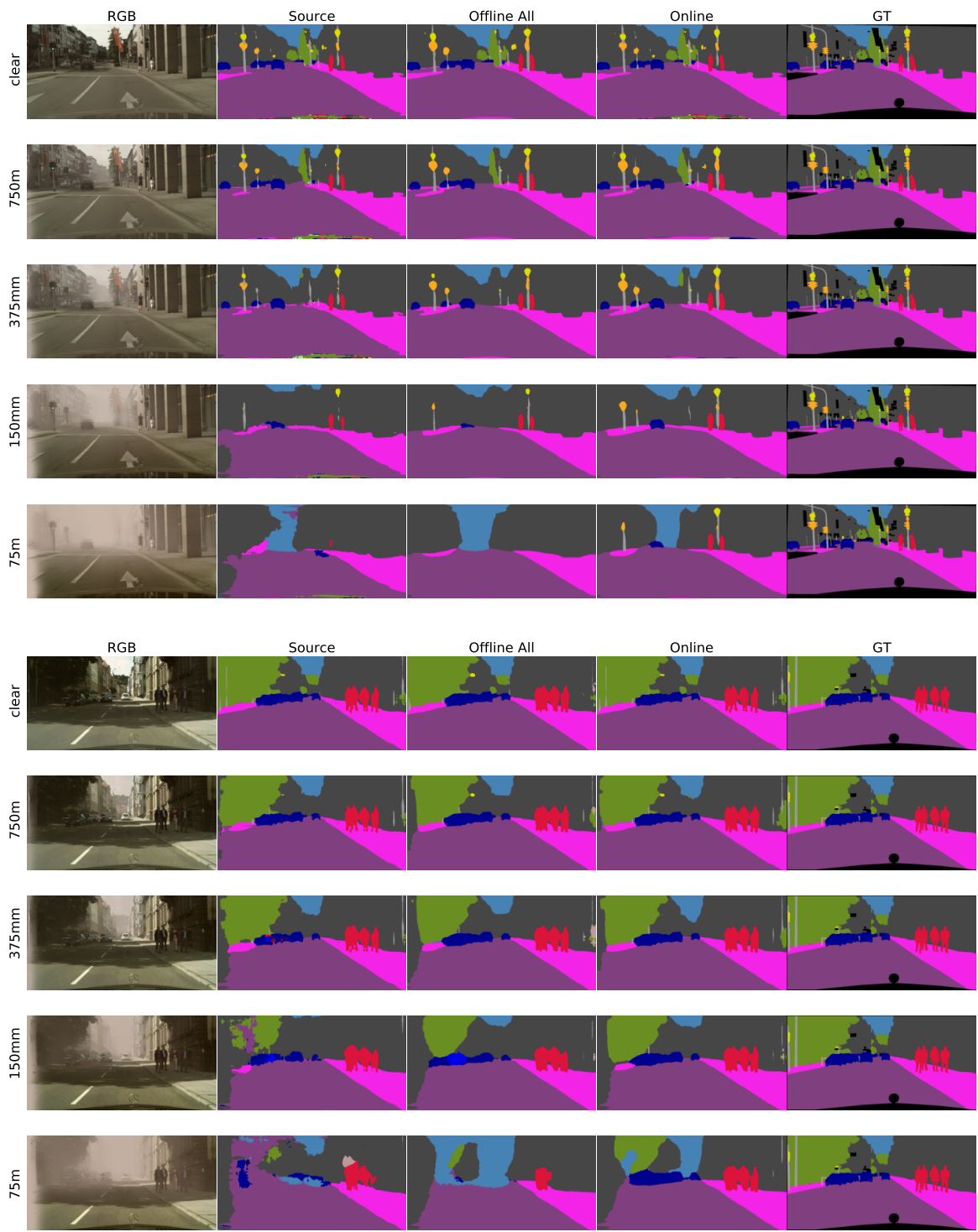

Figure 5. **Qualitative comparison of the Source, Offline and Online Models on foggy images.** From top to bottom, we report a single sample from each domain, from *clear* to 75m visibility.

5

## 2. Implementation Details

We report in detail all the hyper-parameters used to train the described methods. The supervised models were trained until convergence (up to 100 epochs) using a learning rate of $2.5e-4$ and an exponential decay learning rate with a power of $0.9$, together with a weight decay of $5e-4$. For the offline model the same learning policy was used but with an initial learning rate of $1e-5$ and 10 epochs of training. Moreover, for the models using all of the domains (Supervised All & Offline All), a single epoch was considered as passing through all Domains.

The prototypes update was set to $\lambda = 0.9995$ while the momentum model update to $0.999$. The thresholds used for the switches are the following: $T_{cd} = 2e-4$, $T_c = 0.86$, $T_s = 0.94$, $T_d = 0.82$, $T_{c_A} = 0.9$ and $T_{c_B} = 0.83$. Moreover, the Online models were trained using *SGD* with a momentum of $0.9$, learning rates of $8e-4$ and $1e-4$ for the feature extractor and segmentation head respectively, and using a shifting window of length $n = 200$. Lastly, the Advent model was trained using the default settings detailed in [7]. All models performed training with a batch size of $4$ and images scaled to $512 \times 1024$ resolution.

## 3. Domain Shift Detection

As discussed in the main paper, when performing online adaptation it becomes crucial to properly detect the domain shifts and act accordingly. The confidence $h_{\text{static}}$ by the static model can be a good indicator to detect the transition across domains. More precisely, the lower the confidence, the farther the current domain is from the source domain. Then, sudden changes of such a confidence can reflect the transition to a new domain. Moreover, by looking at the sign of such change we can identify if we are moving towards a domain closer to the source one or, vice-versa, if we are going to a new, farther domain. This behavior is encoded by the sign of the derivative of $h_{\text{static}}$.

Figure 6 presents an intuition of this working principle: the blue curve shows the trend of the static model confidence $h_{\text{static}}$, highlighting sudden changes in correspondence of domain changes. The orange curve plots the derivative $\mu_{z_t} - \mu_{z_{t-1}}$ of $h_{\text{static}}$: we can notice how this curve has negative peaks when the confidence drastically drops, i.e. when we approach a new domain farther from the source; on the contrary, we have positive peaks when we switch to a domain closer to the source one.

Based on this principle, we design some of the switching mechanisms introduced and evaluated in our paper.

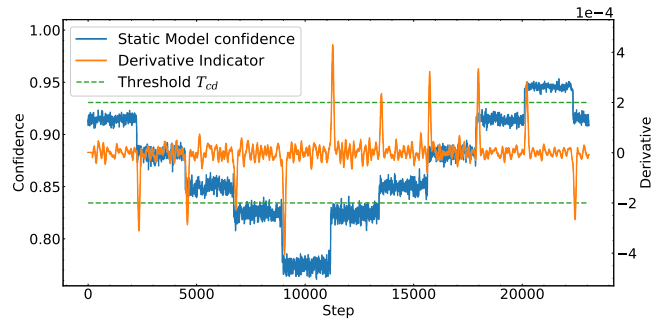

Figure 6. **Working principle of the domain switching detector.** On the back, in blue, the $h_{\text{static}}$ confidence is displayed, while with orange color the derivative $\mu_{z_t} - \mu_{z_{t-1}}$ is showed. When switching to more distant domains the value function has negative peeks and a positive on closer domains.

## 4. Standard Deviation Analysis

Figure 7 displays the progression of the IoU standard deviation across segmentation classes over time. Interestingly, the variance decreases after an adaptation cycle. Before starting adaptation (step 0), we do not observe a specific relationship between the class variance and the domain itself. After a full adaptation cycle, we can notice a correlation between rain intensity and variance – indeed, once the model is back to the *clear* domain, the lower variance (and thus more consistent IoU across all the classes) is achieved on domains closer to the *clear* one, with the highest variance on 200mm (the farthest domain). This hints that the adaptation process itself regularizes the performance across classes, creating an interesting correlation between rain intensity and variance.

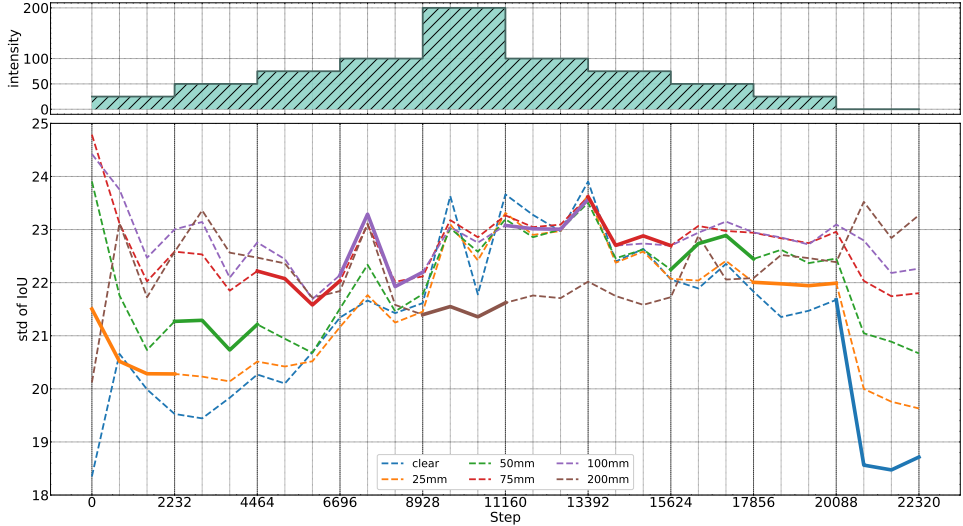

Figure 7. **Illustration of the Standard deviation of the IoU over the classes over time.** Experiment carried out on the Increasing Storm baseline scenario.

## 5. Calibration Error Analysis

Figure 8 shows confidence, mIoU and Expected Calibrtion Error (ECE) [4] of $h_{\text{static}}$ over the target domains (from *clear* to 200mm). Similarly to several other works in Unsupervised Domain Adaption (UDA) and Self-Training, OnDA heavily bases its working principles on model confidence, in particular, on the confidence of the static model, $h_{\text{static}}$. By studying Figure 8 we can notice how the calibration error remains almost constant for domains closer to the source – despite the steady performance drop – proving that the model confidence is correspondingly decreased to match the output accuracy.

This equilibrium does not hold for domains farther from source: indeed, despite confidence remains in general high (∼0.80), the mIoU (and accuracy) dramatically drops, consequently leading to significant increase of the ECE. The analysis of this trend, on one hand, suggests that the confidence of $h_{\text{static}}$ for domains closer to the source represents a valid heuristic. One the other hand, it proves that the static model confidence validity is not universal as its reliability degrades progressively as we move to further domains. We highlight how our Domain Indicator Function, $I_t$ uniquely exploits the confidence derivative (not the absolute value), hence it is not sensitive to model miscalibration as long as the confidence is decreasing with the target domain distance.

## 6. Extended Ablation Study

We conclude this document by digging into three further ablation studies, on learning rate, regularization and Batch Norm respectively.

### 6.1. Learning Rate

According to [5], adapting the weights of the features extractor forces the network to align their representation to those required by the prediction head to achieve good predictions. Inspired by these findings, we investigate the effect of different learning rates applied to the features extractor and the segmentation head separately. Figure 9 shows the behavior of our model while being adapted on the Increasing Storm scenario, focusing respectively on 50mm (left) and 200mm (right) domains. Each curve shows the behavior of one of the two modules (features extractor or prediction head) while varying its learning rate and keeping constant the learning rate for the other module.

Empirically, we found out that setting a higher learning rate for the features extractor (green) allows for better performance, confirming the findings in [5]. Acting on the learning rate of the prediction head (red) shows negligible impact.

### 6.2. Regularization

Table 1 collects results for adaptation on the Increasing Storm scenario, both when proceeding forward (F) or backward (B) across domains. Empirical results show that confidence regularization does not affect performance substantially all-over

7

Figure 8. **Comparison between Confidence, MIoU and Confidence Error (ECE) on the static (source) model across domains.** Experiments carried out on the Increasing Storm scenario. ECE is computed over 1000 bins.

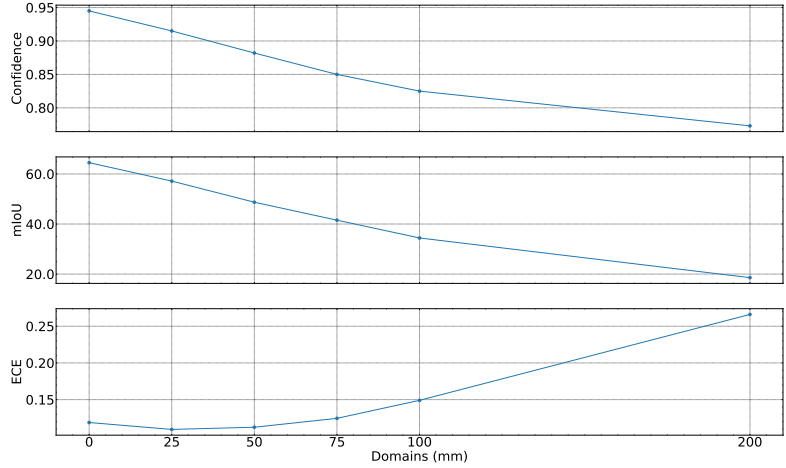

Figure 9. **Learning rate comparison between feature extractor and segmentation head, on the 50mm and 200mm case.** For each testing the learning rate on the opposite sub-network remained fixed at $1e-4$ and $1e-5$ on the feature extractor and segmentation head respectively. Adaptations happens gradually from *clear* to intensities of 25, 50, 75, 100 and 200.

the adaptation process. Nevertheless, by introducing it with a factor 0.1, we are able to increase adaptation performance on the hardest domains, *i.e.* 100mm and $200mm$. On the contrary, regularization itself results less effective when adapting backward, leading to slightly lower mIoU when going back to 50mm, 25mm and *clear* domains.

|  | *clear* | | 25mm | | 50mm | | 75mm | | 100mm | | 200mm | | h mean | |
|---|---|---|---|---|---|---|---|---|---|---|---|---|---|---|
|  | F | B | F | B | F | B | F | B | F | B | F | B | F | B |
| Regularization factor 0   | 64.5 | 65.7 | 59.7 | 59.3 | 54.8 | 55.1 | 51.4 | 50.7 | 48.7 | 45.7 | 38.0 | 38.0 | 51.4 | 50.8 |
| Regularization factor 0.1 | 64.5 | 64.8 | 60.4 | 57.1 | 57.3 | 54.5 | 54.8 | 52.2 | 52.0 | 49.1 | 42.2 | 42.2 | 54.2 | 52.4 |
| Regularization factor 0.2 | 64.5 | 62.6 | 59.3 | 54.2 | 57.4 | 52.4 | 53.9 | 50.5 | 50.2 | 48.4 | 41.4 | 41.4 | 53.4 | 50.8 |
| Regularization factor 0.3 | 64.5 | 60.5 | 59.9 | 54.1 | 56.9 | 52.2 | 52.5 | 50.7 | 50.2 | 49.0 | 41.9 | 41.9 | 53.3 | 50.8 |

Table 1. **Base adaptation cycle results using the Hybrid Switch approach with different confidence regularization factors.** Experiments in the Increasing Storm scenario

## 6.3. Batch Normalization

In Figure 10 we present a comparison between the three Batch-Normalization (BN) policies:

- *BN freeze*: freezing the BN when processing samples from the Replay Buffer,

8

- *BN switching*: swapping BN statistics between target samples and Replay Buffer,
- *Shared BN*: sharing BN statistics.

For simplicity we only present two domains: the source, 0mm (*clear*), and 200mm cases. We underline how these validation metrics are meaningful only when the model is actually exposed to that domain. We can observe that all of the three options perform similarly for the source domain (notice frames 20088 to 22320) on the mIoU (*clear*) plot. However, the *Shared BN* falls behind in terms of adaptation flexibility, as shown in the mIoU (200mm) in the frame interval 8928 to 11160.

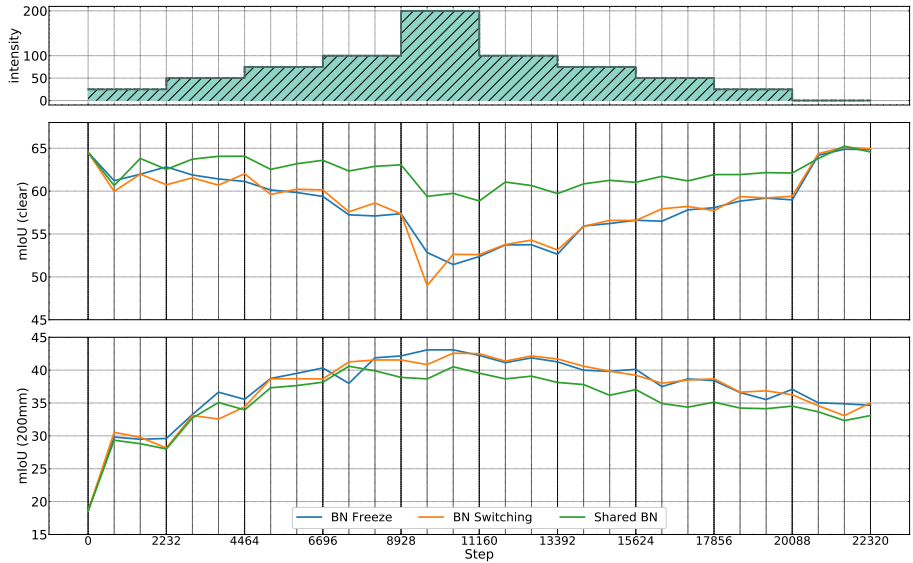

Figure 10. **Comparison between different Batch Normalisation (BN) policies.** The comparison is carried out using the Hybrid Switch on the Increasing Storm baseline scenario.

## References

[1] Jonathan T Barron and Ben Poole. The fast bilateral solver. In *European Conference on Computer Vision*, pages 617–632. Springer, 2016. 1
[2] Marius Cordts, Mohamed Omran, Sebastian Ramos, Timo Rehfeld, Markus Enzweiler, Rodrigo Benenson, Uwe Franke, Stefan Roth, and Bernt Schiele. The cityscapes dataset for semantic urban scene understanding, 2016. 1
[3] Clement Godard, Oisin Mac Aodha, and Gabriel J. Brostow. Unsupervised monocular depth estimation with left-right consistency. In *Proceedings of the IEEE Conference on Computer Vision and Pattern Recognition (CVPR)*, July 2017. 1
[4] Chuan Guo, Geoff Pleiss, Yu Sun, and Kilian Q. Weinberger. On calibration of modern neural networks. In *Proceedings of the 34th International Conference on Machine Learning - Volume 70*, ICML'17, page 1321–1330. JMLR.org, 2017. 7
[5] Jian Liang, Dapeng Hu, and Jiashi Feng. Do we really need to access the source data? source hypothesis transfer for unsupervised domain adaptation. *CoRR*, 2020. 7
[6] Maxime Tremblay, Shirsendu S. Halder, Raoul de Charette, and Jean-François Lalonde. Rain rendering for evaluating and improving robustness to bad weather. *International Journal of Computer Vision*, 2020. 1
[7] Tuan-Hung Vu, Himalaya Jain, Maxime Bucher, Matthieu Cord, and Patrick Pérez. ADVENT: Adversarial entropy minimization for domain adaptation in semantic segmentation. 2019. 6


}

\end{document}